\documentclass[runningheads]{llncs}

\usepackage[mobile]{eccv}

\usepackage{graphicx}
\usepackage{booktabs}

\usepackage[accsupp]{axessibility}  
\usepackage{algorithm}

\usepackage{epsfig}
\usepackage{amsmath}

\usepackage{xcolor,pifont}
\newcommand*\colourcheck[1]{%
  \expandafter\newcommand\csname #1check\endcsname{\textcolor{#1}{\ding{52}}}%
}

\newcommand*\colourxmark[1]{%
  \expandafter\newcommand\csname #1xmark\endcsname{\textcolor{#1}{\ding{56}}}%
}
\colourcheck{green}
\colourxmark{red}

\usepackage{listings}

\usepackage{multirow, makecell, boldline, booktabs}

\usepackage{color, colortbl} 
\definecolor{Gray}{gray}{0.875}
\definecolor{LightCyan}{rgb}{0.9,1,1}

\DeclareMathOperator{\simop}{sim}

\usepackage{algpseudocode}

\usepackage{hyperref}

\usepackage{orcidlink}

\makeatletter
\newcommand{\printfnsymbol}[1]{%
  \textsuperscript{\@fnsymbol{#1}}%
}
\makeatother

\begin{document}

\title{Face Reconstruction Transfer Attack as Out-of-Distribution Generalization} 




\author{Yoon Gyo Jung\thanks{Equal contribution}\inst{1}\orcidlink{0000-0002-0945-4311} \and 
Jaewoo Park\printfnsymbol{1}\inst{2}\orcidlink{0000-0001-7508-3371} \and 
Xingbo Dong\inst{3} \orcidlink{0000-0001-9782-6068} \and
Hojin Park\inst{4} \and \\
Andrew Beng Jin Teoh\inst{5}\orcidlink{0000-0001-5063-9484} \and 
Octavia Camps\inst{1} \orcidlink{0000-0003-1945-9172}
}

\authorrunning{Y.~Jung et al.}

\institute{
Northeastern University ~~
\mbox{\and}AiV Co. ~~
\mbox{\and}Anhui University \and
Hanwha Vision ~~ 
\mbox{\and}Yonsei University 
}

\maketitle

\begin{abstract}
Understanding the vulnerability of face recognition systems to malicious attacks is of critical importance. Previous works have focused on reconstructing face images that can penetrate a targeted verification system. Even in the white-box scenario, however, naively reconstructed images misrepresent the identity information, hence the attacks are easily neutralized once the face system is updated or changed. In this paper, we aim to reconstruct face images which are capable of transferring face attacks on unseen encoders. We term this problem as Face Reconstruction Transfer Attack (FRTA) and show that it can be formulated as an out-of-distribution (OOD) generalization problem. Inspired by its OOD nature, we propose to solve FRTA by Averaged Latent Search and Unsupervised Validation with pseudo target (ALSUV). To strengthen the reconstruction attack on OOD unseen encoders, ALSUV reconstructs the face by searching the latent of amortized generator StyleGAN2 through multiple latent optimization, latent optimization trajectory averaging, and unsupervised validation with a pseudo target. We demonstrate the efficacy and generalization of our method on widely used face datasets, accompanying it with extensive ablation studies and visually, qualitatively, and quantitatively analyses. Code: \url{https://github.com/jungyg/ALSUV.git}
\keywords{Face Reconstruction Transfer Attack \and Face Identity Reconstruction \and Out-of-Distribution Generalization}
\end{abstract}

\begin{figure*}[t]
\centering
\includegraphics[width=.9\linewidth]{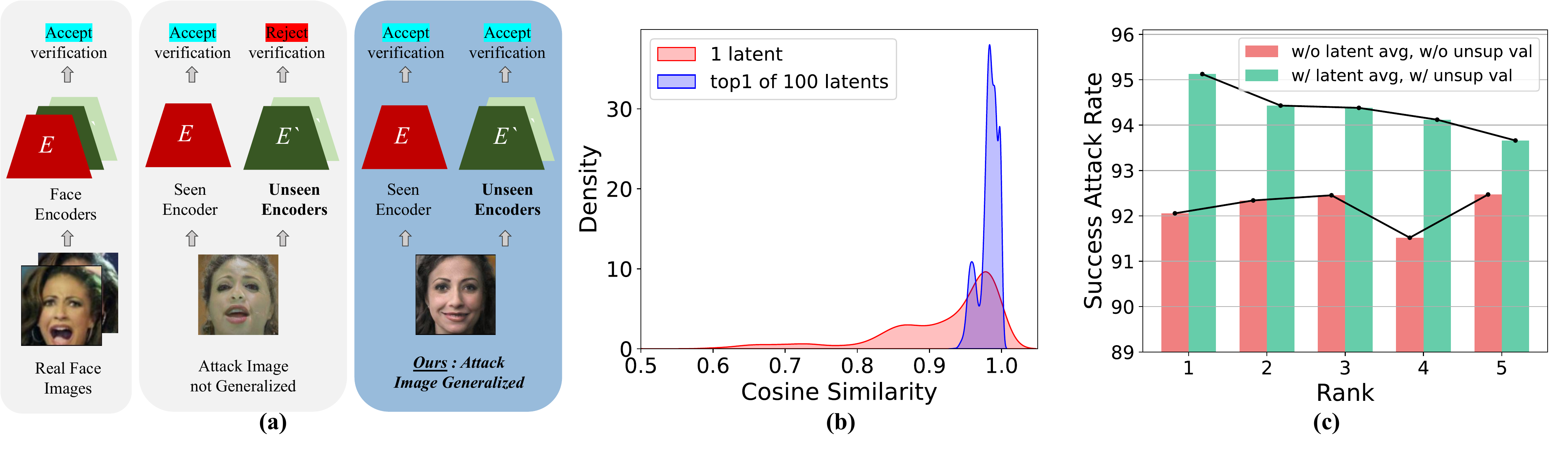}
\caption{
(a) Poorly generalized attack images often get rejected on unseen encoders (middle) while generalized images can bypass other unseen systems (right) like real face images (left).(b) Cosine similarity when only one latent is optimized (red) and the top 1 of multiple latent optimized (blue). Red histogram suffers with underfitting. (c) SAR of top 1 to 5 candidates of target encoder (coral bar) and our method (turquoise bar) tested on unseen encoders. Our method shows consistent correlation between rank and performance as well as better results.
}
\label{fig:problem}
\end{figure*}
%

\section{Introduction}
\label{sec:intro}

    With the increasing deployment of face recognition systems in security-critical environments, threat actors are developing sophisticated attack strategies over various attack points, where one of the major threats is face reconstruction attacks  \cite{mai2018reconstruction,dong2021towards,dong2023reconstruct,razzhigaev2020black,razzhigaev2021darker,duong2020vec2face,shahreza2022face,shahreza2023comprehensive,shahreza2023face}. 
    The primary goal of face reconstruction attacks is to create fake biometric images that resemble genuine ones from the stored biometric templates which are then used to bypass the system. 
    Previous works have mostly focused solely on attacking the target (seen) encoder, i.e., using these fake biometric images to bypass the same system. 
    However, \textit{transfer attack} scenarios, where these fake biometric images are used to bypass other unseen systems (Fig.~\ref{fig:problem}a middle) are not discussed enough. 
    They are potentially more perilous as they can break into a wide range of face recognition systems.
    
    Formally, we define Face Reconstruction Transfer Attacks (FRTA) as successfully reconstructing a face image that can substitute a real face image on unseen encoders, as illustrated in Fig~\ref{fig:problem}a. To state our problem in a rigorous and tractable framework, we formulate this task to reconstruct a face image which \textbf{matches} the original image \textbf{in identity} by a finite number of unseen face encoders given only a single encoder and a feature embedded through this encoder (section \ref{sec.formalization}). However, existing works pay less attention to transfer attacks. 

\begin{figure}[t]
\centering
\includegraphics[width=.995\linewidth]{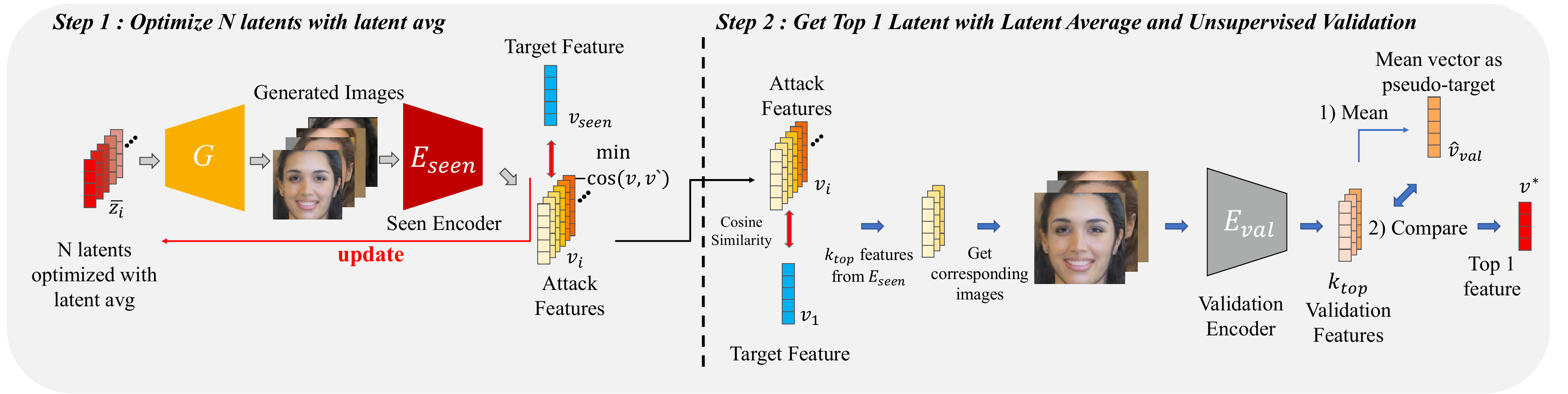}
\caption{Overview of our method. Latents of a pre-trained generative model $G$ is optimized to ensure high similarity between the feature embedding of reconstructed samples and real feature embeddings.}
\label{fig:method}
\end{figure}

    To solve FRTA effectively, we first devise a novel out-of-distribution (OOD) oriented FRTA framework that reformulates the attack as a problem of generalization of loss function over OOD of network parameters. 
    Our FRTA task falls to the standard OOD generalization category in that the loss function needs to be optimized by one variable and generalized with respect to the other instantiated from unseen distributions. 
    We postulate a white-box scenario\footnote{The experiment in black-box scenario is given in Sec.~\ref{sec:alsuv_black_box}} where a single encoder and a template feature embedded through this encoder are given, and face reconstruction is achieved by optimizing the data input in a way to minimize feature reconstruction loss. 
    Instead of directly updating the input image, we adopt a generative model and update its latent which outputs an image (Fig.~\ref{fig:method} step 1).

    However, reconstructing the face with naive latent optimization is likely to suffer from  underfitting with poor latent optimization (Fig.~\ref{fig:problem}b red histogram).
    To address this challenge, we introduce an Averaged Latent Search with Unsupervised Validation through pseudo target (ALSUV) framework, which is motivated by the OOD generalization concept. In this OOD-oriented approach, we 1) optimize multiple latents concurrently by 2) employing latent averaging and 3) searching for the most optimal generalized sample through unsupervised validation using a validation encoder. 
    Multiple latent optimization prevents poor optimization by selecting the well-optimized sample close to the target (Fig.~\ref{fig:problem}b blue histogram). However, this potentially causes overfitting to the seen encoder (Fig.~\ref{fig:problem}c coral bars). 
    Therefore, we average latents throughout optimization trajectories which provides a flatter loss surface (Fig.~\ref{fig:loss_surf}) and leads to OOD generalization \cite{petzka2021relative,cha2021swad,rame2022diverse}. 
    Additionally, we adopt an unsupervised validation search with pseudo target, which incorporates a surrogate validation encoder to search for the best generalizing sample in validation encoder space. 
    Overall, latent averaging and unsupervised validation alleviate overfitting to seen encoder as shown in Fig.~\ref{fig:problem}c (turquoise bar) where rank and performance are notably correlated.
    The contributions of our work are summarized as:
\begin{itemize}


\item We rigorously formulate the reconstruction transfer attack as an OOD generalization problem. Inspired by this framing, we effectively improve the transfer attack performance by an OOD-oriented method ALSUV.

\item Our method achieves state-of-the-art in the transfer attack scenarios based on LFW, CFP-FP, and AgeDB with 6 different types of face encoders under the measures of success attack rate and top-1 identification rate.


\item We conduct in-depth empirical analysis on the components of ALSUV.
Particularly, we observe that ALSUV improves the transfer attack performance by finding a flatter minima in the reconstruction loss landscape, evidencing the close connection between face transfer attack and OOD generalization.

\end{itemize}

\section{Related Works}
    \subsection{Face Reconstruction from Features} 
    NBNet \cite{mai2018reconstruction} first pioneered face reconstruction from template by neighborly deconvolution, but the results are substandard for both quality and performance. \cite{dong2021towards} projects features into the latent space of a pre-trained StyleGAN2 \cite{karras2019style} to generate fine-grained resembling images. The results are qualitatively decent, but often contain different identities as shown in Fig.~\ref{fig:sample_images}. DiBiGAN \cite{duong2020vec2face} presents a generative framework based on bijective metric learning and pairs features with face images one-to-one. These methods offer fast sample generation after training but require extensive face datasets and time for training a new network. \cite{razzhigaev2020black} samples varying random Gaussian blobs iteratively and combining the blobs as the shape of a face. It requires only a few queries and no prior knowledge such as dataset, but shows results with low quality. \cite{razzhigaev2021darker} reconstructs faces using similarity scores based on eigenface with soft symmetry constraints, generative regularization, and multi-start policy to avoid local minimas. However, the reconstructed images show severe noise as shown in Fig.~\ref{fig:sample_images}. These works consume less time to generate few samples than training a new network but takes longer for large scale generations.
    
    Methods introduced above show promising results when tested with seen encoders, but performance drastically falls with unseen encoders. Recently, \cite{dong2023reconstruct,park2023towards,shahreza2023face,shahreza2023comprehensive} regarded transfer attack scenarios in their works. \cite{dong2023reconstruct} suggests a genetic algorithm-based approach along with an attack pipeline to impersonate the target user. Reconstructed images are high quality, but evolutionary algorithms relying on random mutation and selection processes might get trapped in local optima or fail to generalize well due to the limited exploration-exploitation trade-off in their inherent design. \cite{park2023towards} suggests query efficient zeroth-order gradient estimation with ensembling the initial top $k$ initialization search followed by ensembling. This approach iteratively adjusts latent representations to minimize reconstruction errors but suffer from overfitting to specific characteristics of the seen encoder. 
    \cite{shahreza2023face,shahreza2023comprehensive} train networks which map face embeddings to the $\mathcal{W^{+}}$ latent space of StyleGAN3 and 3D aware generative model respectively to utilize the representation of each generative models. They evaluate 5 different types of attacks including transfer attack in whitebox and blackbox scenario in their work. Unlike previous works, we approach FRTA as an OOD generalization problem with more challenging datasets and various types of face recognition systems.


    \subsection{Out-of-Distribution Generalization} Generalization is one of the most important tasks in deep learning models especially when it comes to unseen OOD circumstances. Searching for flat minima is one of the main stems of research to achieve generalization where \cite{petzka2021relative} establishes a strong connection between the flatness of the loss surface and generalization in deep neural networks. Weight averaging\cite{garipov2018loss,izmailov2018averaging,cha2021swad,rame2022diverse,petzka2021relative} ensembles the trajectories of non-linear function parameters during training to seek well generalized flat minima point. \cite{garipov2018loss} gathers information from separately trained models, \cite{izmailov2018averaging} stochastically averages a single model with cyclic learning rate, \cite{cha2021swad} aggregates from a dense trajectory, and \cite{rame2022diverse} collects from several different training policies. Our task focuses on generalizing reconstructed face over unseen encoders, hence, we adopt the core principles from these works and adequately modify them for our work. 
    
    Pseudo label has been used for generalization tasks where label information is scarce such as domain adaptation\cite{saito2018maximum,shin2020two,zou2018unsupervised}. They can be made by mixing both samples and labels\cite{zhang2017mixup}, ensembling labels from augmented samples \cite{berthelot2019mixmatch,hu2021simple}, or confidence prediction with sliding window voting followed by confidence-based prediction\cite{shin2020two}. The key components for pseudo labels are sufficiently high confidence \cite{hu2021simple,shin2020two} and adequate regularization to prevent over-confidence\cite{zou2018unsupervised,zou2019confidence}. Our method requires selecting the best generalizing input(latent) in an unsupervised manner. Hence, we migrate the idea of previous works and discover how to find the proper pseudo target for our task.

\section{Face Reconstruction Transfer Attack as OOD Generalization\label{sec.formalization}}
We first rigorously define the problem of FRTA, and we show that face reconstruction transfer attack is in fact an OOD generalization problem.

\subsection{Problem Formalization}
The FRTA can be formalized as an algorithm $\mathcal{A}$ that must return a solution vector $\mathcal{A}(\theta_{seen}) = x^*$ to the optimization objective
\begin{equation}
\label{eq:obj}
\max_x \min_{\theta \in \Theta} sim(E_{\theta}(x), v_\theta),
\end{equation}
where $v_\theta$ indicates a true target $v_\theta=E_{\theta}(x_{real})$ corresponding to an encoder $E_{\theta}$. The nature of FRTA poses a constraint that the attacker can access to the seen encoder $E_{\theta_{seen}}$ only. The encoder parameter space $\Theta$ includes both seen and unseen encoder networks exposed to attacks. Hence, the objective requires that the attack must be \textit{transferrable}.

One approach to the attack is by using a face image generator $G$. By substituting $x=G(z)$ in the above equation, the objective is maximized in terms of $z$ instead of $x$: 
$
\max_z \min_{\theta \in \Theta} sim(E_{\theta}(G(z)), v_\theta).
$

\subsection{FRTA as OOD Generalization}
Given that both $E_{\theta}$ and $G$ are multi-layer perceptron instances, we show that the FRTA can be formulated as an OOD generalization problem. To this end, we first formalize the OOD generalization as follows:

\noindent
\textbf{Definition.}
OOD generalization on the domain $\mathcal{D}$ in the parameter space $\Theta$ is to solve for an algorithm $\mathcal{A}$ that, given a seen dataset, returns a solution parameter $\mathcal{A}(D_{seen}) = \theta^*$ that minimizes
\begin{equation}
\min_{\theta} \max_{D \in \mathcal{D}} L(\theta; D),
\end{equation}
where $L$ is a loss function defined as
\begin{equation}
L(\theta; D) = \frac{1}{\left| D \right|}
\sum_{(x,y) \in D} 
l(f_\theta(x), y).
\end{equation}
Here, $f_\theta$ is an MLP parametrized by $\theta$, and $l$ is a sample-wise loss function on a label-data pair. 
\qed

\noindent
\textbf{Theorem.}
Define $f_z$ by $f_z(\theta) = E_\theta( G(z))$,
and let $D^*_{seen}=\{ (\theta_{seen}, v_{\theta_{seen}}) \}$, $\mathcal{D}^* = \{ \{ (\theta, v_{\theta}) \} : \theta \in \Theta \}$, and
\begin{equation}
l(f_z(\theta), v_\theta) = - sim(f_z(\theta), v_\theta).
\end{equation}
Then, $f_z$ is an MLP, and the FRTA algorithm $\mathcal{A}$ on $\Theta$ is an OOD generalization algorithm $\mathcal{A}^*$ on the domain $\mathcal{D}^*$ in the parameter space $\mathcal{Z}$.

The theorem is proved (in Supp.~\ref{sec:supp_proof}) by observing the duality between data and parameter; in MLP, data can be viewed as parameter, and vice versa. 

\subsection{Averaged Latent Search with Unsupervised Validation with Pseudo Target(ALSUV)}
Inspired by the above interpretation, we tackle FRTA by means of OOD generalization techniques by defining the similarity as a loss function. Thereupon, we propose ALSUV with pseudo target, which is an integrated approach of OOD generalization on the latent. 

The latent search mechanism of ALSUV are decomposed as follows: (1) multiple latent optimization, (2) latent averaging throughout optimization trajectories, and (3) unsupervised validation with the pseudo target. 

\subsubsection{Multiple Latent Optimization}
In order to avoid the underminimization problem shown in Fig.~\ref{fig:problem}b and generate candidates for our following unsupervised validation method, we initialize multiple $n$ latent vectors and optimize them in a parallel manner:
\begin{equation}
\label{eq:plo}
\underset{ \{z_i\}_{i=1}^n }{\min}
\,
\sum_{i=1}^n L( z_i; E_{seen} )= -\sum_{i=1}^{n}sim(E_{seen}(G(z_{i})), v_{seen})
\end{equation}
where $E_{seen} = E_{\theta_{seen}}$ $v_{seen} = v_{\theta_{seen}}$. 
The given loss function is minimized by a gradient-based update using the standard optimizer such as Adam\cite{kingma2014adam} or SGD.
Fig.~\ref{fig:ablation_graphs} validates that updating with multiple latents significantly improves the minimization of the loss. 
Moreover, this method can effectively escape from poor local minima than iterating with complicated learning rate scheduler (Tab.~\ref{table:ablation_periodic}).

\subsubsection{Latent Averaging}
Avoiding the poor underminimization issue alone is not sufficient for effective generalization of attack under FRTA since the acquired solution may overfit to seen encoders as shown in Fig.~\ref{fig:problem}c. 
To improve the attack rate on the unseen encoders, we borrow the idea from OOD generalization\cite{petzka2021relative,cha2021swad,rame2022diverse} and apply averaging the solution latent vectors over the optimization trajectory: 
\begin{equation}
\label{eq:la}
    \overline{z}_i = \frac{1}{t_{avg}} \sum_{t=T-t_{avg}}^{T} z^{(t)}_i
\end{equation}
where $z^{(t)}_i$ are the latent vectors acquired at step $t$ of the optimization of Eq. \eqref{eq:plo}, $T$ is the total number of optimization steps, and $t_{avg}$ is the size of steps to average the latent vectors.

Due to the equivalence between FRTA and OOD generalization from the theorem above, latent averaging can improve the generalization of the similarity maximization under unseen encoder networks and corresponding unseen targets. 
Fig.~\ref{fig:ablation_graphs} evidences our hypothesis, indicating that averaging the latents improves the attack rate on the unseen encoders. 
Moreover, Fig.~\ref{fig:loss_surf} shows that latent averaging improves the transfer attack by smoothening the loss surface of our objective, validating the equivalence between OOD generalization and FRTA.

\subsubsection{Unsupervised Validation with Pseudo Target} 
Despite its effectiveness, multiple latent optimization with averaging can still suffer overfitting issues as no explicit information of unseen encoders is exposed to the optimization. Therefore, we acquire more explicit information of unseen encoders by utilizing a surrogate validation encoder $E_{val} = E_{\theta_{val}}$. 
Particularly, we propose to validate our reconstruction criterion Eq.~\eqref{eq:obj} over the surrogate validation encoder. 
One remaining issue however is the absence of validation target vector $v_{val}$. To resolve it, we construct the pseudo target: 
\begin{equation}
\label{eq:uv}
\widehat{v}_{val}= \frac{1}{k_{top}} \sum_{i=1}^{k_{top}} E_{\theta_{val}}(G(\overline{z}_{(i)})),
\end{equation}
by averaging the reconstructed features from the top $k$ latent vectors of the attack objective in Eq.~\eqref{eq:plo}. Namely,  $\overline{z}_{(i)}$ are ordered with respect to the similarity to the seen feature $L(\overline{z}_{(1)}) \leq \dots \leq L(\overline{z}_{(n)})$. The pseudo target may not be fully precise approximation of validation target $v_{val}$ from the real ground truth image. However, we show that it improves the attack under FRTA by mitigating the overfitting issue of the latent optimization (Fig.~\ref{fig:ablation_graphs}c). Moreover,  we find that the average of multiple top $k$ reconstructed features better serves as an alternative of $v_{val}$ than the single top $1$ feature as the latter may have been overfitted to the seen target.

\subsubsection{Full Objective}
Overall, ALSUV searches the solution latent $z^*$ by unsupervised validation
\begin{equation}
z^* = \arg\min_{\overline{z}_i}
d(E_{val}(G( \overline{z}_{i})), \widehat{v}_{val}),
\end{equation}
subject to $\overline{z}_i \in S$
within the search space $S$ of multiple latent-averaged vectors defined in Eqs.~\eqref{eq:plo} and \eqref{eq:la}
based on a pseudo target $\widehat{v}_{val}$ defined in Eq.~\eqref{eq:uv}. The reconstruction target feature of ALSUV avoids under-minimization of latent optimization, thereby effectively attacking the seen encoder. On the other hand, it achieves robust FRTA based on latent averaging and validation against pseudo target with the surrogate validation encoder. The full algorithm is given in Supp.~\ref{sec:supp_method}.

\section{Experiments}
The experiments section includes: 1) performance evaluation against existing methods; 2) comprehensive component ablation and hyperparameter variation to show effectiveness; 
3) component-wise in-depth ablation study and analysis on ALSUV.

    
    \subsection{Configuration}
    We use StyleGAN2\cite{karras2019style} trained on FFHQ-256 for the generator $G$, and the latents are optimized in the $\mathcal{W^{+}}$ space. Both $G$ and target encoders $E_{seen}$ are frozen while optimizing latents. We adopt Adam\cite{kingma2014adam} optimizer with 100 steps where the learning rate starts from 0.1 and is divided by 10 at iteration 50. Our method involves three hyperparameters: $n=100$, the number of latents; $t_{avg}=70$, length of trajectory for latent averaging; and $k_{top}=10$, the number of samples used for unsupervised validation. We use pytorch\cite{paszke2019pytorch} for all experiments on a single Nvidia RTX 2080ti GPU.   

\begin{table*}[t]
    \caption{SAR of previous works and our proposed method. Red cells indicate results of seen encoder setup. We highlight our method with grey and best results are highlighted with bold.}
    \centering
    \resizebox{0.97\linewidth}{!}{
    \begin{tabular}{ll|cccccc|c|cccccc|c|cccccc|c}
    \toprule 
         \multicolumn{2}{c|}{Dataset} &  
         \multicolumn{7}{c|}{\textbf{LFW}} & 
         \multicolumn{7}{c|}{\textbf{CFP-FP}} & 
         \multicolumn{7}{c}{\textbf{AgeDB-30}} \\
        \midrule
         \multicolumn{2}{c|}{Test Encoder}  & 
         FaceNet & MobFace & ResNet50 & ResNet100 & Swin-S & VGGNet & Unseen AVG & 
         FaceNet & MobFace & ResNet50 & ResNet100 & Swin-S & VGGNet & Unseen AVG & 
         FaceNet & MobFace & ResNet50 & ResNet100 & Swin-S & VGGNet & Unseen AVG \\ 
        \midrule
        Target Encoder & \scshape{Real Face} & 
        99.87 & 99.83 & 99.83 & 99.93 & 99.93 & 99.8 & - & 
        93.85 & 76.15 & 95.38 & 95.38 & 90 & 93.08 & - & 
        92.63 & 93.53 & 92.34 & 97.17 & 96.23 & 91.7 & - \\ 
        \midrule
        
        \multirow{8}{*}{FaceNet\cite{schroff2015facenet}} & \scshape{NBNet} & 
        82.91\cellcolor{red!25} & 0.57 & 70.58 & 34.85 & 0.74 & 74.15 & 36.51 & 
        80.77 \cellcolor{red!25}& 25.38 & 73.85 & 36.15 & 18.46 & 73.08 & 47.69 & 
        69.91 \cellcolor{red!25}& 2.57 & 63.69 & 19.57 & 2.7 & 60.99 & 32.49 \\ 
         & \scshape{LatentMap} & 
         60.28 \cellcolor{red!25}& 5.33 & 53.76 & 4.52 & 8.4 & 59.47 & 31.74 & 
         49.23 \cellcolor{red!25}& 21.54 & 36.15 & 13.08 & 44.62 & 41.54 & 35.96 & 
         49.53\cellcolor{red!25} & 8.05 & 35.37 & 1.61 & 3.51 & 39.97 & 21.73 \\ 
         & \scshape{Genetic} & 
         87.72\cellcolor{red!25} & 8.5 & 63.52 & 7.46 & 16.1 & 76.04 & 34.32 & 
         80.77\cellcolor{red!25} & 26.92 & 58.46 & 25.38 & 39.23 & 54.62 & 40.92 & 
         66.82\cellcolor{red!25} & 9.04 & 46.31 & 7.05 & 7.98 & 49.76 & 24.03 \\ 
         & \scshape{GaussBlob} & 
         0.4\cellcolor{red!25} & 0.2 & 0.03 & 0 & 0 & 2.8 & 0.61 & 
         9.23\cellcolor{red!25} & 6.92 & 10 & 1.54 & 6.15 & 9.23 & 6.77 & 
         6.15\cellcolor{red!25} & 3.7 & 11.59 & 0.39 & 1.32 & 14.29 & 6.26 \\    
         & \scshape{EigenFace} & 
         91.46\cellcolor{red!25} & 17.82 & 56.67 & 24.6 & 21.77 & 69.59 & 38.09 & 
         84.62\cellcolor{red!25} & 26.92 & 58.46 & 19.23 & 33.85 & 60 & 39.69 & 
         73.87\cellcolor{red!25} & 9.49 & 53.14 & 8.79 & 3.96 & 55.71 & 26.22 \\ 
         & \scshape{FaceTI} & 
         3.4\cellcolor{red!25} & 2.06 & 4.15 & 0.3 & 0.57 & 8.46 & 3.11 & 
         13.08\cellcolor{red!25} & 6.92 & 10 & 1.54 & 6.15 & 9.23 & 6.77 & 
         16.96\cellcolor{red!25} & 3.7 & 11.59 & 0.39 & 1.32 & 14.29 & 6.26 \\ 
         & \scshape{QEZOGE} & 
         99.6\cellcolor{red!25} & 25.95 & 93.29 & 59.72 & 67.85 & 96.39 & 68.64 & 
         91.54\cellcolor{red!25} & 39.23 & 83.85 & 58.46 & 67.69 & 77.69 & 65.38 & 
         88.12\cellcolor{red!25} & 19.99 & 72.42 & 31.09 & 24.59 & 79.08 & 45.43 \\ 
         \rowcolor{Gray} \cellcolor{white} & \scshape{Ours} &  
         \textbf{99.83} & \textbf{65.15} & \textbf{97.51} & \textbf{86.79} & \textbf{83.32} & \textbf{99.22} & \textbf{86.4} & 
         \textbf{94.62} & \textbf{46.15} & \textbf{86.15} & \textbf{73.08} & \textbf{73.85} & \textbf{86.15} & \textbf{73.08} & 
         \textbf{89.67} & \textbf{30.8} & \textbf{79.18} & \textbf{48.89} & \textbf{38.46} & \textbf{83.39} & \textbf{56.14} \\
        \midrule
        
        \multirow{8}{*}{MobFaceNet\cite{chen2018mobilefacenets,wang2020mis}} & \scshape{NBNet} & 
        4.52 & 30.17\cellcolor{red!25} & 2.56 & 2.36 & 3.44 & 7.41 & 4.06 & 
        12.31 & 49.23\cellcolor{red!25} & 18.46 & 9.23 & 16.15 & 10 & 13.23 & 
        13.32 & 21.15\cellcolor{red!25} & 12.58 & 4.06 & 2.32 & 16.7 & 9.8 \\ 
         & \scshape{LatentMap} & 
         4.18 & 7.12\cellcolor{red!25} & 7.96 & 0.88 & 2.29 & 8.61 & 4.78 & 
         13.85 & 14.62\cellcolor{red!25} & 17.69 & 1.54 & 12.31 & 15.38 & 12.15 & 
         11.91 & 6.63\cellcolor{red!25} & 12.42 & 0.74 & 1.64 & 14.23 & 8.19 \\ 
         & \scshape{Genetic} & 
         13.87 & 49.54\cellcolor{red!25} & 15.05 & 4.42 & 8.17 & 18.8 & 12.06 & 
         23.08 & 45.38\cellcolor{red!25} & 26.92 & 13.85 & 28.46 & 28.46 & 24.15 & 
         19.31 & 33.02\cellcolor{red!25} & 19.76 & 3.57 & 5.31 & 20.86 & 13.76 \\ 
         & \scshape{GaussBlob} & 
         2.97 & 75.57\cellcolor{red!25} & 3.37 & 4.62 & 6.61 & 7.29 & 4.97 & 
         11.54 & 61.54\cellcolor{red!25} & 13.85 & 3.85 & 10.77 & 13.85 & 10.77 & 
         11.43 & 45.25\cellcolor{red!25} & 8.24 & 6.28 & 4.63 & 13.81 & 8.88 \\  
         & \scshape{EigenFace} &  
         62.44 & 99.39\cellcolor{red!25} & 63.45 & 56.87 & 75.26 & 65.54 & 64.71 & 
         51.54 & 70.77\cellcolor{red!25} & 59.23 & 47.69 & 64.62 & 54.62 & 55.54 & 
         59.51 & 88.86\cellcolor{red!25} & 51.66 & 35.66 & 40.88 & 56.32 & 48.81 \\
         & \scshape{FaceTI} &  
         15.67 & 21.91\cellcolor{red!25} & 18.2 & 4.55 & 9.81 & 24.5 & 14.55 & 
         19.23 & 23.08\cellcolor{red!25} & 23.85 & 6.92 & 13.85 & 21.54 & 17.08 & 
         29.51 & 14\cellcolor{red!25} & 24.46 & 3.35 & 4.6 & 28.52 & 18.09 \\
         & \scshape{QEZOGE} &  
         61.51 & 97.98\cellcolor{red!25} & 67.81 & 54.36 & 65.28 & 69.87 & 63.77 & 
         39.23 & 70.77\cellcolor{red!25} & 44.62 & 32.31 & 49.23 & 43.08 & 41.69 & 
         50.98 & 79.76\cellcolor{red!25} & 52.04 & 30.35 & 34.15 & 56 & 44.7 \\
         \rowcolor{Gray} \cellcolor{white} & \scshape{Ours} &
         \textbf{96.76} & \textbf{99.83} & \textbf{97.3} & \textbf{96.83} & \textbf{99.16} & \textbf{98.05} & \textbf{97.62} & 
         \textbf{80} & \textbf{79.23} & \textbf{81.54} & \textbf{84.62} & \textbf{83.85} & \textbf{83.85} & \textbf{82.77} & 
         \textbf{85.1} & \textbf{93.72} & \textbf{84.68} & \textbf{87.93} & \textbf{87.58} & \textbf{86.22} & \textbf{86.3} \\
        \midrule
        
        \multirow{8}{*}{ResNet50\cite{he2016deep,wang2018cosface}} & \scshape{NBNet} & 
        61.09 & 1.79 & 84.68\cellcolor{red!25} & 34.12 & 0.3 & 78.1 & 35.08 & 
        64.62 & 23.85 & 77.69\cellcolor{red!25} & 46.92 & 14.62 & 66.92 & 43.39 & 
        68.62 & 4.31 & 76.67\cellcolor{red!25} & 31.12 & 1.71 & 72.55 & 35.66 \\ 
         & \scshape{LatentMap} & 
         24.97 & 5.7 & 29.67\cellcolor{red!25} & 2.36 & 9.11 & 33.68 & 15.16 & 
         33.08 & 16.15 & 28.46\cellcolor{red!25} & 7.69 & 19.23 & 30.77 & 21.38 & 
         30.25 & 4.57 & 29.61\cellcolor{red!25} & 1.71 & 2.22 & 30.96 & 13.94 \\ 
         & \scshape{Genetic} & 
         61.09 & 12.55 & 83.46\cellcolor{red!25} & 8.2 & 13.7 & 72.53 & 33.61 & 
         49.23 & 19.23 & 60.77\cellcolor{red!25} & 20.77 & 40 & 49.23 & 35.69 & 
         46.64 & 10.04 & 60.32\cellcolor{red!25} & 7.76 & 8.34 & 50.02 & 24.56 \\ 
         & \scshape{GaussBlob} & 
         0.03 & 0.1 & 0.2\cellcolor{red!25} & 0 & 0 & 0.4 & 0.11 & 
         3.08 & 22.31 & 6.15\cellcolor{red!25} & 1.54 & 2.31 & 6.15 & 7.08 & 
         2.67 & 1.32 & 4.18\cellcolor{red!25} & 0.29 & 0.29 & 4.92 & 1.9 \\  
         & \scshape{EigenFace} & 
         78.87 & 35.3 & 95.75\cellcolor{red!25} & 47.69 & 48.8 & 84.48 & 59.03 & 
         70 & 38.46 & 81.54\cellcolor{red!25} & 51.54 & 45.38 & 78.46 & 56.77 & 
         69.04 & 19.83 & 81.14\cellcolor{red!25} & 25.1 & 26.62 & 73.32 & 42.78 \\ 
         & \scshape{FaceTI} & 
         12.87 & 8.93 & 18.77\cellcolor{red!25} & 1.58 & 4.31 & 21.27 & 9.79 & 
         20.77 & 13.08 & 23.08\cellcolor{red!25} & 6.15 & 16.15 & 21.54 & 15.54 & 
         32.99 & 11.3 & 28.23\cellcolor{red!25} & 2.45 & 3.54 & 31.61 & 16.38 \\
         & \scshape{QEZOGE} & 
         86.01 & 42.97 & 98.31\cellcolor{red!25} & 68.32 & 64.34 & 90.06 & 70.34 & 
         73.85 & 39.23 & 86.92\cellcolor{red!25} & 53.85 & 63.08 & 73.85 & 60.77 & 
         71.26 & 20.05 & 85.48\cellcolor{red!25} & 35.66 & 27.39 & 77.02 & 46.28 \\
         \rowcolor{Gray} \cellcolor{white}& \scshape{Ours} &     
         \textbf{99.12} & \textbf{85.24} & \textbf{99.8} & \textbf{97.44} & \textbf{97.61} & \textbf{99.63} & \textbf{95.81} & 
         \textbf{89.23} & \textbf{65.38} & \textbf{95.38} & \textbf{82.31} & \textbf{84.62} & \textbf{86.92} & \textbf{81.69} & 
         \textbf{88.86} & \textbf{56.2} & \textbf{91.6} & \textbf{76.6} & \textbf{71.32} & \textbf{89.28} & \textbf{76.45} \\
        \midrule

        \multirow{8}{*}{ResNet100\cite{he2016deep,deng2019arcface}} & \scshape{NBNet} & 
        20.18 & 4.08 & 28.35 & 48.36\cellcolor{red!25} & 8.98 & 30.1 & 18.34 & 
        33.08 & 27.69 & 41.54 & 51.54\cellcolor{red!25} & 28.46 & 46.15 & 35.38 & 
        37.66 & 8.43 & 48.41 & 62.92\cellcolor{red!25} & 16.83 & 47.67 & 31.8 \\
         & \scshape{LatentMap} & 
         6.68 & 3.27 & 7.69 & 3.04\cellcolor{red!25} & 3.61 & 11.81 & 6.61 & 
         20 & 15.38 & 16.92 & 12.31\cellcolor{red!25} & 21.54 & 23.08 & 19.38 & 
         6.57 & 3.22 & 8.24 & 1.09\cellcolor{red!25} & 0.68 & 9.56 & 5.654 \\ 
         & \scshape{Genetic} & 
         23.42 & 9.55 & 25.31 & 47.76\cellcolor{red!25} & 15.22 & 29.97 & 20.69 &
         41.54 & 28.46 & 35.38 & 48.46\cellcolor{red!25} & 41.54 & 33.08 & 36 
         & 23.79 & 9.69 & 21.98 & 38.56\cellcolor{red!25} & 10.75 & 24.2 & 18.08 \\ 
         & \scshape{GaussBlob} & 
         31.7 & 51.86 & 60.7 & 85.52\cellcolor{red!25} & 71.3 & 48.72 & 52.86 & 
         35.38 & 21.54 & 43.85 & 61.54\cellcolor{red!25} & 45.38 & 40.77 & 37.38 &
         50 & 35.83 & 52.29 & 73.63\cellcolor{red!25} & 50.76 & 54.32 & 48.64 \\ 
         & \scshape{EigenFace} & 
         85.69 & 69.96 & 87.61 & 95.41\cellcolor{red!25} & 91.33 & 87.24 & 84.37 &
         65.38 & 45.38 & 63.85 & 85.38\cellcolor{red!25} & 78.46 & 63.08 & 63.23 &
         79.11 & 52.4 & 81.88 & 93.24\cellcolor{red!25} & 81.2 & 80.08 & 74.93 \\ 
         & \scshape{FaceTI} & 
         15.57 & 38.22 & 21.47 & 39.33\cellcolor{red!25} & 29.29 & 28.95 & 26.7 & 
         28.46 & 36.92 & 42.31 & 45.38\cellcolor{red!25} & 50.77 & 40.77 & 39.85 & 
         20.12 & 29.87 & 26.39 & 37.88\cellcolor{red!25} & 23.46 & 31.19 & 26.21 \\     
         & \scshape{QEZOGE} & 
         55.88 & 48.57 & 69.94 & 98.89\cellcolor{red!25} & 73.78 & 71.39 & 63.91 & 
         53.08 & 40.77 & 53.85 & 86.15\cellcolor{red!25} & 64.62 & 55.38 & 53.54 & 
         47.51 & 25.59 & 50.53 & 91.37 \cellcolor{red!25} & 45.38 & 54.84 & 44.77 \\ 
         \rowcolor{Gray} \cellcolor{white} & Ours & 
         \textbf{99.36} & \textbf{96.16} & \textbf{99.56} & \textbf{99.93} & \textbf{99.93} & \textbf{99.56} & \textbf{98.91} & 
         \textbf{91.54} & \textbf{67.69} & \textbf{95.38} & \textbf{95.38} & \textbf{90} & \textbf{90} & \textbf{86.92} & 
         \textbf{92.11} & \textbf{85.07} & \textbf{91.73} & \textbf{97.01} & \textbf{94.85} & \textbf{91.76} & \textbf{91.1} \\
        \midrule
        
        \multirow{8}{*}{Swin-S\cite{liu2021swin,wang2020mis}} & \scshape{NBNet} &
        12.76 & 7.19 & 15.79 & 16.67 & 9.65\cellcolor{red!25} & 16.4 & 13.76 & 
        23.85 & 26.15 & 35.38 & 29.23 & 30.77\cellcolor{red!25} & 28.46 & 28.61 & 
        41.65 & 11.01 & 44 & 36.34 & 25.01\cellcolor{red!25} & 43.16 & 35.23 \\ 
         & \scshape{LatentMap} & 
         3.61 & 1.08 & 3.88 & 0.4 & 1.28\cellcolor{red!25} & 6.14 & 3.02 & 
         16.15 & 16.15 & 15.38 & 3.08 & 19.23\cellcolor{red!25} & 16.92 & 13.54 &
         8.27 & 2.32 & 8.72 & 0.9 & 1.19\cellcolor{red!25} & 9.85 & 6.01 \\ 
         & \scshape{Genetic} & 
         35.61 & 21.03 & 39.39 & 12.66 & 51.3\cellcolor{red!25} & 45.9 & 30.92 & 
         45.38 & 33.08 & 37.69 & 29.23 & 65.38\cellcolor{red!25} & 35.38 & 36.15 & 
         30.87 & 12.78 & 28.26 & 11.39 & 26.52\cellcolor{red!25} & 29.42 & 22.54 \\ 
         & \scshape{GaussBlob} & 
         0.2 & 3.51 & 0.37 & 1.21 & 5.4\cellcolor{red!25} & 0.34 & 1.13 & 
         10 & 27.69 & 11.54 & 5.38 & 20\cellcolor{red!25} & 6.92 & 12.31 & 
         5.28 & 7.82 & 8.72 & 4.22 & 6.12\cellcolor{red!25} & 9.72 & 7.15 \\ 
         & \scshape{EigenFace} &
         17.48 & 50.46 & 33.24 & 30.88 & 56.94\cellcolor{red!25} & 65.49 & 56.45 & 
         35.38 & 34.62 & 36.15 & 20.77 & 50\cellcolor{red!25} & 43.85 & 44.31 & 
         40.88 & 29.39 & 42.26 & 16.99 & 39.97\cellcolor{red!25} & 38.91 & 33.69 \\
         & \scshape{FaceTI} &
         51.9 & 65.15 & 59.82 & 39.87 & 61.61 \cellcolor{red!25} & 37.9 & 33.99 &
         43.08 & 46.15 & 44.62 & 43.85 & 58.46 \cellcolor{red!25} & 33.08 & 32 &
         37.59 & 30.13 & 36.21 & 20.6 & 23.82 \cellcolor{red!25} & 39.27 & 32.76 \\ 
         & \scshape{QEZOGE} &
         78.7 & 66.94 & 84.33 & 80.59 & 97.44 \cellcolor{red!25} & 86.52 & 79.42 & 
         67.69 & 50 & 69.23 & 65.38 & 81.54 \cellcolor{red!25} & 72.31 & 64.92 & 
         59.54 & 32.76 & 57.45 & 55.17 & 83.13 \cellcolor{red!25} & 61.15 & 53.21 \\
         \rowcolor{Gray} \cellcolor{white} & \scshape{Ours} &    
         \textbf{98.08} & \textbf{99.02} & \textbf{99.02} & \textbf{99.87} & \textbf{99.93} & \textbf{99.36} & \textbf{99.07} & 
         \textbf{87.69} & \textbf{73.08} & \textbf{89.23} & \textbf{93.08} & \textbf{90.77} & \textbf{88.46} & \textbf{86.31} & 
         \textbf{89.35} & \textbf{87.19} & \textbf{89.15} & \textbf{95.72} & \textbf{96.3} & \textbf{90.15} & \textbf{90.31} \\ 
        \midrule
        
        \multirow{8}{*}{VGGNet\cite{simonyan2014very,wang2018cosface}} & \scshape{NBNet} & 
        45.57 & 1.38 & 63.5 & 16.51 & 2.39 & 73.78\cellcolor{red!25} & 25.87 & 
        56.15 & 22.31 & 65.38 & 37.69 & 24.62 & 69.23\cellcolor{red!25} & 41.23 & 
        63.05 & 3.89 & 67.46 & 21.63 & 4.25 & 70.49\cellcolor{red!25} & 32.06 \\ 
         & \scshape{LatentMap} & 
         21.13 & 3.78 & 32.3 & 2.5 & 2.77 & 38.34\cellcolor{red!25} & 12.5 &
         28.46 & 13.08 & 26.15 & 7.69 & 20 & 28.46\cellcolor{red!25} & 19.08 &
         22.01 & 4.8 & 23.88 & 0.74 & 1.51 & 29.77\cellcolor{red!25} & 10.59 \\ 
         & \scshape{Genetic} & 
         60.01 & 9.21 & 66.05 & 9.58 & 13.36 & 85.29\cellcolor{red!25} & 31.64 & 
         50 & 21.54 & 52.31 & 21.54 & 33.08 & 61.54\cellcolor{red!25} & 35.69 & 
         45.38 & 7.53 & 40.49 & 3.7 & 4.67 & 55.75\cellcolor{red!25} & 20.35 \\ 
         & \scshape{GaussBlob} &
         0.4 & 0.07 & 0.3 & 0 & 0 & 3\cellcolor{red!25} & 0.15 & 
         0.67 & 21.54 & 6.92 & 0.77 & 2.31 & 8.46\cellcolor{red!25} & 6.44 & 
         6.03 & 0.92 & 4.53 & 0.2 & 0 & 11.74\cellcolor{red!25} & 2.34 \\ 
         & \scshape{EigenFace} & 
         93.82 & 25.99 & 93.93 & 44.52 & 57.21 & 99.12\cellcolor{red!25} & 63.09 & 
         84.62 & 37.69 & 86.15 & 62.31 & 60 & 89.23\cellcolor{red!25} & 66.15 &
         72.67 & 14.55 & 72.55 & 18.44 & 15.45 & 81.2\cellcolor{red!25} & 38.73 \\ 
         & \scshape{FaceTI} & 
         2.36 & 2.46 & 2.46 & 0.3 & 1.28 & 7.41 \cellcolor{red!25} & 1.77 & 
         7.69 & 6.15 & 9.23 & 0.77 & 7.69 & 13.85 \cellcolor{red!25} & 6.31 & 
         25.04 & 6.76 & 17.25 & 1.09 & 3.06 & 23.98 \cellcolor{red!25} & 10.64 \\
         & \scshape{QEZOGE} & 
         90.26 & 47.32 & 91.64 & 64.54 & 60.57 & 98.69 \cellcolor{red!25} & 70.87 & 
         77.69 & 43.85 & 79.23 & 60 & 65.38 & 88.46 \cellcolor{red!25} & 65.23 & 
         73.9 & 19.76 & 75.12 & 28.71 & 25.43 & 85.9 \cellcolor{red!25} & 44.58 \\

         \rowcolor{Gray} \cellcolor{white} & \scshape{Ours} & 
         \textbf{99.29} & \textbf{77.45} & \textbf{99.49} & \textbf{94.57} & \textbf{93.9} & \textbf{99.8} & \textbf{92.94} & 
         \textbf{89.23} & \textbf{63.85} & \textbf{90.77} & \textbf{87.69} & \textbf{85.38} & \textbf{91.54} & \textbf{83.38} & 
         \textbf{83.97} & \textbf{37.4} & \textbf{83.55} & \textbf{55.62} & \textbf{47.38} & \textbf{89.25} & \textbf{61.58} \\
    \bottomrule
    \end{tabular}
    }

\label{table:result_sar}
\end{table*}

\begin{table*}[t]
    \caption{
    Rank 1 identification rate of previous works and our proposed method. Red cells indicate results of seen encoder setup. We highlight our method with grey and best results are highlighted with bold.
    }
    \centering
    \renewcommand{\arraystretch}{1.1}
    \resizebox{0.97\linewidth}{!}{
    \begin{tabular}{ll|cccccc|c|cccccc|c|cccccc|c}
    \toprule 
          \multicolumn{2}{c|}{Datsaet} &  
          \multicolumn{7}{c|}{\textbf{LFW}} & 
          \multicolumn{7}{c|}{\textbf{CFP-FP}} & 
          \multicolumn{7}{c}{\textbf{AgeDB-30}}  \\
        \midrule
            \multicolumn{2}{c|}{Test Encoder} & 
            FaceNet & MobFace & ResNet50 & ResNet100 & Swin-S & VGGNet & Unseen AVG & 
            FaceNet & MobFace & ResNet50 & ResNet100 & Swin-S & VGGNet & Unseen AVG & 
            FaceNet & MobFace & ResNet50 & ResNet100 & Swin-S & VGGNet & Unseen AVG \\ 
        \midrule
           Target Encoder & \scshape{Real Face} & 
           99 & 99.5 & 99.5 & 99.5 & 99.5 & 99.5 & - & 
           95.5 & 88.5 & 93.5 & 100 & 99 & 94 & - & 
           92 & 96 & 92.5 & 98 & 98 & 87.5 & - \\ 
        \midrule
        \multirow{8}{*}{FaceNet} & \scshape{NBNet} & 
        86.07\cellcolor{red!25} & 1 & 57.71 & 39.3 & 3.48 & 53.23 & 30.94 & 
        62 \cellcolor{red!25} & 1.0 & 46.5 & 38.5 & 5 & 50 & 28.3 &
        92.5\cellcolor{red!25} & 0 & 56 & 21.5 & 3.5 & 60 & 28.2 \\
         & \scshape{LatentMap} & 
         32.5\cellcolor{red!25} & 2 & 20 & 12.5 & 11.5 & 19.5 & 13.1 & 
         17.5\cellcolor{red!25} & 1.5 & 14 & 10 & 8 & 13 & 9.3 &
         15.5\cellcolor{red!25} & 1.5 & 7 & 2 & 2 & 7.5 & 4 \\ 
         & \scshape{Genetic} & 
         82.5\cellcolor{red!25} & 7.5 & 39 & 18.5 & 23 & 39 & 25.4 & 
         18 \cellcolor{red!25} & 3 & 15.5 & 11.5 & 10.5 & 18 & 11.7 &
         43\cellcolor{red!25} & 1.5 & 14.5 & 6.5 & 4 & 17.5 & 8.8 \\  
         & \scshape{GaussBlob} & 
         0.5\cellcolor{red!25} & 0.5 & 0 & 0 & 0 & 0 & 0.1 & 
         1 \cellcolor{red!25} & 0.5 & 1 & 1 & 0 & 0.5 & 0.6 &
         0.5\cellcolor{red!25} & 0 & 0 & 0 & 0 & 0 & 0 \\ 
         & \scshape{EigenFace} & 
         91.54\cellcolor{red!25} & 13.93 & 48.26 & 30.35 & 17.91 & 51.24 & 32.34 & 
         72 \cellcolor{red!25} & 2 & 39.5 & 17.5 & 8 & 34 & 20.2 &
         69.5\cellcolor{red!25} & 3 & 23 & 6.5 & 7.5 & 30.5 & 14.1 \\ 
         & \scshape{FaceTI} & 
         1 \cellcolor{red!25} & 0 & 0 & 1 & 0.5 & 2.49 & 0.8 & 
         0.5 \cellcolor{red!25} & 0.5 & 1.5 & 0 & 0 & 0.5 & 0.5 & 
         0.5 \cellcolor{red!25} & 0 & 0.5 & 0.5 & 0.5 & 0.5 & 0.4 \\
         & \scshape{QEZOGE} & 
         100\cellcolor{red!25} & 25.37 & 90.55 & 77.61 & 69.15 & 94.03 & 71.34 & 
         91 \cellcolor{red!25} & 16 & 66 & 65 & 52 & 73.5 & 54.5 & 
         96 \cellcolor{red!25} & 9.5 & 59.5 & 33 & 26 & 70 & 39.6 \\
         \rowcolor{Gray} \cellcolor{white} & \scshape{Ours} & 
         \textbf{100} & \textbf{55.5} & \textbf{100} & \textbf{94.53} & \textbf{91} & \textbf{100} & \textbf{88.21} & 
         \textbf{95.5} & \textbf{21} & \textbf{79.5} & \textbf{81} & \textbf{63.5} & \textbf{85} & \textbf{66} & 
         \textbf{98} & \textbf{21.5} & \textbf{80.5} & \textbf{56} & \textbf{45.5} & \textbf{87.5} & \textbf{58.2} \\
        \midrule
        
        \multirow{8}{*}{MobFace} & \scshape{NBNet} & 
        1.49 & 47.76\cellcolor{red!25} & 1 & 12.44 & 8.45 & 4.98 & 5.67 & 
        0.5 & 4.5\cellcolor{red!25} & 2.5 & 5 & 2 & 2.5 & 2.5 &
        2 & 41.5\cellcolor{red!25} & 2 & 13 & 7.5 & 3.5 & 5.6 \\ 
         & \scshape{LatentMap} & 
         3.48 & 4.48\cellcolor{red!25} & 2.49 & 1.99 & 1.49 & 1.49 & 2.19 & 
         3 & 1\cellcolor{red!25} & 1.5 & 1 & 3 & 2.5 & 2.2 &
         0 & 3.5\cellcolor{red!25} & 1 & 0.5 & 1 & 2 & 0.9 \\ 
         & \scshape{Genetic} & 
         2.99 & 68.66\cellcolor{red!25} & 7.46 & 6.97 & 9.95 & 6.47 & 6.77 & 
         6.5 & 5\cellcolor{red!25} & 3 & 3.5 & 5 & 5.5 & 4.7 &
         2.5 & 53\cellcolor{red!25} & 5 & 3.5 & 2.5 & 7 & 4.1 \\ 
         & \scshape{GaussBlob} &
         0 & 89.05\cellcolor{red!25} & 2.99 & 25.87 & 21.89 & 3.98 & 10.95 & 
         3 & 10\cellcolor{red!25} & 1.5 & 6 & 2.5 & 2 & 3 &
         1.5 & 81.5\cellcolor{red!25} & 1.5 & 16 & 15.5 & 1.5 & 7.2 \\ 
         & \scshape{EigenFace} & 
         46.27 & 100\cellcolor{red!25} & 55.22 & 74.13 & 90.05 & 52.74 & 63.68 & 
         30.5 & 76\cellcolor{red!25} & 37.5 & 46 & 56.5 & 33.5 & 40.8 &
         33 & 99.5\cellcolor{red!25} & 38 & 54 & 77.5 & 43 & 49.1 \\     
         & \scshape{FaceTI} & 
         10.45 & 17.41\cellcolor{red!25} & 8.96 & 9.45 & 12.44 & 8.96 & 10.05 & 
         7.5 & 8 \cellcolor{red!25} & 5.5 & 8.5 & 7 & 6 & 6.9 & 
         4 & 7.5 \cellcolor{red!25} & 4 & 3.5 & 4 & 2.5 & 3.6 \\
         & \scshape{QEZOGE} & 
         38.81 & 100 \cellcolor{red!25} & 47.76 & 59.7 & 65.67 & 45.77 & 51.54 & 
         22 & 60.5 \cellcolor{red!25} & 27.5 & 46.5 & 40.5 & 25.5 & 32.4 & 
         24 & 99.5 \cellcolor{red!25} & 28.5 & 43 & 50 & 29.5 & 35 \\
         
         \rowcolor{Gray} \cellcolor{white} & \scshape{Ours} & 
         \textbf{94.5} & \textbf{100} & \textbf{99} & \textbf{99.5} & \textbf{100} & \textbf{98.5} & \textbf{98.3} & 
         \textbf{75.5} & \textbf{87} & \textbf{78.5} & \textbf{92.5} & \textbf{94} & \textbf{78} & \textbf{83.7} & 
         \textbf{92.5} & \textbf{99.5} & \textbf{97} & \textbf{97} & \textbf{98.5} & \textbf{94} & \textbf{95.8} \\
        \midrule
        
        \multirow{8}{*}{ResNet50} & \scshape{NBNet} & 
        60.5 & 0.5 & 88.5\cellcolor{red!25} & 53.5 & 3 & 60.5 & 35.6 & 
        48 & 0.5 & 61 \cellcolor{red!25} & 43 & 3.5 & 50 & 29 &
        65 & 0.5 & 95\cellcolor{red!25} & 40.5 & 5.5 & 76 & 37.5 \\
         & \scshape{LatentMap} & 
         9 & 2.5 & 12.5\cellcolor{red!25} & 2.36 & 9.5 & 14.5 & 7.57 & 
         9 & 2 & 11.5 \cellcolor{red!25} & 11.5 & 7 & 10.5 & 8 &
         5 & 2 & 13.5\cellcolor{red!25} & 3 & 3 & 8 & 4.2 \\ 
         & \scshape{Genetic} & 
         29 & 5.5 & 80.5\cellcolor{red!25} & 18 & 16.5 & 47 & 23.2 & 
         16 & 3 & 23.5 \cellcolor{red!25} & 14.5 & 11 & 12 & 11.3 &
         15.5 & 4 & 65.5\cellcolor{red!25} & 5 & 7.5 & 23.5 & 11.1 \\ 
         & \scshape{GaussBlob} & 
         0 & 0 & 0\cellcolor{red!25} & 0 & 0 & 0 & 0 & 
         0.5 & 0 & 0 \cellcolor{red!25} & 0 & 0.5 & 0.5 & 0.3 &
         0 & 0.5 & 0\cellcolor{red!25} & 0.5 & 1 & 0.5 & 0.5 \\ 
         & \scshape{EigenFace} & 
         69.15 & 29.35 & 99\cellcolor{red!25} & 65.67 & 67.16 & 81.59 & 62.58 & 
         50 & 7 & 74.5 \cellcolor{red!25} & 49.5 & 32 & 60.5 & 39.8 &
         55 & 24 & 99\cellcolor{red!25} & 43 & 45 & 79 & 49.2 \\        
         & \scshape{FaceTI} & 
         5.47 & 3.48 & 6.97 \cellcolor{red!25} & 3.48 & 4.98 & 6.47 & 4.78 & 
         4 & 2.5 & 5.5  \cellcolor{red!25} & 5.5 & 6 & 4.5 & 4.5 & 
         3.5 & 1.5 & 6 \cellcolor{red!25} & 3 & 1 & 5 & 2.8 \\ 
         & \scshape{QEZOGE} & 
         79.6 & 31.84 & 100 \cellcolor{red!25} & 75.12 & 67.16 & 88.06 & 68.36 & 
         63.5 & 19.5 & 83 \cellcolor{red!25} & 64 & 53.5 & 64.5 & 53 & 
         66.5 & 12.5 & 100 \cellcolor{red!25} & 47 & 31 & 82.5 & 47.9 \\   
         \rowcolor{Gray} \cellcolor{white} & \scshape{Ours} & 
         \textbf{100} & \textbf{90} & \textbf{100} & \textbf{100} & \textbf{99} & \textbf{100} & \textbf{97.8} &
         \textbf{91.5} & \textbf{52} & \textbf{94} & \textbf{95.5} & \textbf{86.5} & \textbf{89} & \textbf{82.9} & 
         \textbf{98} & \textbf{74} & \textbf{100} & \textbf{89.5} & \textbf{85} & \textbf{99.5} & \textbf{89.2} \\
        \midrule
        
        \multirow{8}{*}{ResNet100} & \scshape{NBNet} & 
        21 & 7 & 31 & 91\cellcolor{red!25} & 26.5 & 20 & 21.1 & 
        11 & 0.5 & 19.5 & 57.5 \cellcolor{red!25} & 11 & 19 & 12.2 &
        18.5 & 2.5 & 24 & 91\cellcolor{red!25} & 33 & 18 & 19.2 \\ 
         & \scshape{LatentMap} & 
         3.48 & 1.49 & 2.49 & 6.47\cellcolor{red!25} & 4.98 & 2.99 & 3.09 & 
         6 & 2.5 & 5.5 & 11.5 \cellcolor{red!25} & 7.5 & 7 & 5.7 &
         1.5 & 0 & 0.5 & 1.5\cellcolor{red!25} & 0 & 0 & 0.4 \\ 
         & \scshape{Genetic} & 
         14.43 & 6.97 & 12.94 & 75.12\cellcolor{red!25} & 25.37 & 14.93 & 14.93 & 
         7.5 & 4 & 6 & 24\cellcolor{red!25} & 8.5 & 6.5 & 6.5 &
         4 & 0.5 & 3 & 70.5\cellcolor{red!25} & 8.5 & 4 & 4 \\ 
         & \scshape{GaussBlob} & 
         34.5 & 52 & 51.5 & 90\cellcolor{red!25} & 75.5 & 46 & 51.9 & 
         17.5 & 8.5 & 29.5 & 70.5 \cellcolor{red!25} & 38.5 & 23.5 & 23.5 &
         20.19 & 39.42 & 32.69 & 89.42\cellcolor{red!25} & 71.15 & 25 & 37.69 \\ 
         & \scshape{EigenFace} & 
         76.12 & 79.6 & 85.07 & 99.5\cellcolor{red!25} & 96.52 & 78.11 & 83.08 & 
         54.5 & 23.5 & 65.5 & 95.5 \cellcolor{red!25} & 85 & 56.5 & 57 &
         57 & 67 & 66.5 & 98.5\cellcolor{red!25} & 96 & 61.5 & 69.6 \\       
         & \scshape{FaceTI} & 
         16.42 & 35.32 & 19.9 & 70.65 \cellcolor{red!25} & 39.3 & 16.92 & 25.57 & 
         16 & 24.5 & 18 & 55 \cellcolor{red!25} & 32 & 19 & 21.9 & 
         7.5 & 12 & 7.5 & 63 \cellcolor{red!25} & 22.5 & 7 & 11.3 \\
         & \scshape{QEZOGE} & 
         50.75 & 34.83 & 48.26 & 100 \cellcolor{red!25} & 71.14 & 51.74 & 51.34 & 
         44 & 17 & 49.5 & 97 \cellcolor{red!25} & 61 & 47 & 43.7 & 
         16.5 & 10.5 & 27 & 99 \cellcolor{red!25} & 52.5 & 19 & 25.1 \\         
         \rowcolor{Gray} \cellcolor{white} & \scshape{Ours} & 
         \textbf{99.5} & \textbf{99.5} & \textbf{99.5} & \textbf{100} & \textbf{100} & \textbf{98.51} & \textbf{99.4} & 
         \textbf{89} & \textbf{70.5} & \textbf{92} & \textbf{100} & \textbf{98} & \textbf{90.5} & \textbf{88} & 
         \textbf{86} & \textbf{86} & \textbf{91.5} & \textbf{100} & \textbf{99} & \textbf{88.5} & \textbf{90.2} \\
        \midrule
        
        \multirow{8}{*}{Swin-S} & \scshape{NBNet} & 
        14.5 & 8 & 20 & 47 & 39.5\cellcolor{red!25} & 15.5 & 21 & 
        12.5 & 1 & 14.5 & 29 & 13 \cellcolor{red!25} & 10.5 & 13.5 &
        20.5 & 6 & 24 & 57.5 & 58.5\cellcolor{red!25} & 16 & 24.8 \\ 
         & \scshape{LatentMap} & 
         2 & 0.5 & 3.5 & 1.5 & 1.5\cellcolor{red!25} & 3.5 & 2.2 & 
         3.5 & 1.5 & 2.5 & 6 & 5 \cellcolor{red!25} & 3.5 & 3.4 &
         0 & 0 & 0 & 0 & 2.5\cellcolor{red!25} & 0 & 0 \\ 
         & \scshape{Genetic} & 
         15.5 & 9 & 17 & 27.5 & 67\cellcolor{red!25} & 19 & 17.6 & 
         9 & 3 & 7.5 & 9.5 & 18 \cellcolor{red!25} & 8 & 7.4 &
         6.5 & 4.5 & 4.5 & 13.5 & 52.49\cellcolor{red!25} & 4 & 6.6 \\ 
         & \scshape{GaussBlob} & 
         0 & 3.98 & 0 & 7.46 & 22.89\cellcolor{red!25} & 0 & 2.29 & 
         0 & 1.5 & 1 & 4 & 6.5 \cellcolor{red!25} & 0.5 & 1.4 &
         1 & 5.5 & 1 & 10.5 & 16.5\cellcolor{red!25} & 1.5 & 3.9 \\ 
         & \scshape{EigenFace} & 
         9.95 & 59.2 & 17.41 & 46.27 & 75.62\cellcolor{red!25} & 16.92 & 29.95 & 
         7.5 & 12.5 & 13.5 & 22.5 & 40.5 \cellcolor{red!25} & 13 & 13.8 &
         8 & 45 & 13.5 & 34.5 & 69.5\cellcolor{red!25} & 13.5 & 22.9 \\         
         & \scshape{FaceTI} & 
         35.82 & 55.22 & 36.32 & 50.25 & 63.18 \cellcolor{red!25} & 34.83 & 42.49 & 
         28 & 30.5 & 32 & 44 & 46 \cellcolor{red!25} & 27.5 & 32.4 & 
         13 & 20 & 8.5 & 21 & 31 \cellcolor{red!25} & 11.5 & 14.8 \\         
         & \scshape{QEZOGE} & 
         57.71 & 57.21 & 67.16 & 87.56 & 100 \cellcolor{red!25} & 64.68 & 66.86 & 
         44 & 29.5 & 49 & 78.5 & 90  \cellcolor{red!25} & 52 & 50.6 & 
         23 & 20 & 28 & 68 & 95.5 \cellcolor{red!25} & 18.5 & 31.5 \\
         \rowcolor{Gray} \cellcolor{white} & \scshape{Ours} & 
         \textbf{98.5} & \textbf{100} & \textbf{100} & \textbf{100} & \textbf{100} & \textbf{99.5} & \textbf{99.6} & 
         \textbf{80} & \textbf{77.5} & \textbf{83} & \textbf{99.5} & \textbf{99} & \textbf{83} & \textbf{84.6} & 
         \textbf{80.5} & \textbf{94.5} & \textbf{89.5} & \textbf{99.5} & \textbf{100} & \textbf{86.5} & \textbf{90.1} \\
        \midrule
        
        \multirow{8}{*}{VGGNet} & \scshape{NBNet} & 
        48.76 & 1.99 & 59.7 & 38.81 & 9.95 & 70.15\cellcolor{red!25} & 31.84 &
        45.5 & 1 & 49 & 36.5 & 8.5 & \cellcolor{red!25} 48.5 & 28.1 &
        61.5 & 0.5 & 71.5 & 35.5 & 7 & 88.5\cellcolor{red!25} & 35.2 \\ 
         & \scshape{LatentMap} & 
         7 & 1.5 & 9 & 5 & 4 & 11\cellcolor{red!25} & 5.3 & 
         7.5 & 0.5 & 9 & 3.5 & 4.5 & 10 \cellcolor{red!25} & 5 &
         3.5 & 0.5 & 5.5 & 0.5 & 0.5 & 3.5\cellcolor{red!25} & 2.1 \\ 
         & \scshape{Genetic} & 
         37.5 & 6 & 43 & 21.5 & 16 & 71.5\cellcolor{red!25} & 24.8 & 
         16.5 & 1 & 16.5 & 10 & 9.5 & 21.5 \cellcolor{red!25} & 10.7 &
         17 & 1.5 & 19 & 4 & 4 & 42\cellcolor{red!25} & 9.1 \\ 
         & \scshape{GaussBlob} & 
         0 & 0 & 0 & 0 & 0 & 1.99\cellcolor{red!25} & 0 & 
         0.5 & 0 & 1 & 0 & 0 & 1.5 \cellcolor{red!25} & 0.3 &
         0.51 & 0 & 1.02 & 0 & 0 & 1.53\cellcolor{red!25} & 0.306 \\ 
         & \scshape{EigenFace} & 
         92.54 & 36.32 & 92.54 & 73.13 & 73.13 & 99.5\cellcolor{red!25} & 73.53 & 
         74.5 & 11 & 73.5 & 64 & 48.5 & 88.5 \cellcolor{red!25} & 54.3 &
         74 & 15.5 & 80.5 & 29 & 27 & 93.5\cellcolor{red!25} & 45.2 \\        
         & \scshape{FaceTI} & 
         1.49 & 2.49 & 2.99 & 1.99 & 1.49 & 3.98 \cellcolor{red!25} & 2.09 & 
         3 & 2 & 4.5 & 2 & 2 & 4.5 \cellcolor{red!25} & 2.7 & 
         1.5 & 0.5 & 1.5 & 0.5 & 1 & 1.5 \cellcolor{red!25} & 1 \\
         & \scshape{QEZOGE} & 
         91.54 & 39.3 & 93.53 & 76.12 & 75.12 & 99 \cellcolor{red!25} & 75.12 & 
         72.5 & 16.5 & 72 & 70 & 55 & 84.5  \cellcolor{red!25} & 57.2 & 
         72 & 13 & 76 & 35.5 & 26 & 96 \cellcolor{red!25} & 44.5 \\
         \rowcolor{Gray} \cellcolor{white} & \scshape{Ours}  & 
         \textbf{99.5} & \textbf{82} & \textbf{100} & \textbf{99} & \textbf{99} & \textbf{100} & \textbf{95.9} & 
         \textbf{90} & \textbf{34} & \textbf{88.5} & \textbf{93} & \textbf{84.5} & \textbf{93.5} & \textbf{78} & 
         \textbf{92.5} & \textbf{43} & \textbf{93.5} & \textbf{68} & \textbf{64.5} & \textbf{98.5} & \textbf{72.3} \\
    \bottomrule
    \end{tabular}
    }
\label{table:result_iden}
\end{table*}

    \subsection{Datasets and Networks}
    We use LFW, CFP-FP, and AgeDB-30, three widely used verification datasets with distinct characters. 
    We randomly sample 200 non-overlapping identities each from 3 datasets. 
    Encoders are selected with distinct backbones equipped with various classification heads trained with distinct datasets. We use pre-trained models from \cite{wang2021facex} and models trained by our own trained with CASIA-WebFace\cite{yi2014learning} or MS-1M\cite{Guo2016MSCeleb1MAD}. Specific configurations of dataset and encoders are shown in Supp.~\ref{sec:supp_setup}. For the validation encoder, we use Swin-T as default.

    \subsection{Evaluation Metrics and Details}
    \cite{mai2018reconstruction} introduces Type I and Type II SAR where Type I compares the generated face with the ground truth target image, while Type II compares with different images with the same identity. SAR measures the ratio of generated samples passing the positive verification test where thresholds are specific to type of datasets and face encoders. Since Type I is relatively easy, we only report Type II performance which is more challenging. We also evaluate identification rate, which retrieves the top 1 sample from a gallery set composed of real images with probes consisting of generated samples. We include target samples in the gallery. 
    
    In ablation, we also report the SAR@FAR which is the success attack rate at thresholds of FAR($=1e-4, 1e-3, 1-e2$).
    
    \subsection{Comparison with Previous Works\label{sec.compare2previous}}
    We compare our method with state-of-the-art feature-based face reconstruction methods including NBNet \cite{mai2018reconstruction}, LatentMap \cite{dong2021towards}, Genetic \cite{dong2023reconstruct}, GaussBlob \cite{razzhigaev2020black}, Eigenface \cite{razzhigaev2021darker}, FaceTI \cite{shahreza2023face}, and QEZOGE \cite{park2023towards}. For FaceTI \cite{shahreza2023face}, we used StyleGAN2 and our face encoders for reproduction. As shown in Tab.~\ref{table:result_sar} and Tab.~\ref{table:result_iden}, overall previous methods are effective on seen encoders, but the performance drastically drops on unseen encoders. EigenFace \cite{razzhigaev2021darker}, FaceTI \cite{shahreza2023face}, and QEZOGE \cite{park2023towards} show better performance compared with other works, however, tend to show lower performance compared with our method and the results highly fluctuates depending on the type of seen encoder. In contrast, our method outperforms for both seen and unseen cases. Our SAR and identification rate results are close to real face images on seen encoders while \textbf{outperforming previous works with a large margin on unseen encoders for every dataset while depending less on the type of seen encoder}. Additionally, results tested on unseen encoders shown in Tab.~\ref{table:result_sar} and Tab.~\ref{table:result_iden} present that our method achieves OOD generalization on unseen encoders successfully. We additionally attack SOTA face recognition systems AdaFace \cite{Kim_2022_CVPR}, CurricularFace \cite{Huang_2020_CVPR}, and ElasticFace \cite{Boutros_2022_CVPR} in Tab.~\ref{table:result_sota_fr_systems}.

\begin{table*}[t]
    \caption{
    Ablation study controlling the components of our method. Performance is averaged over the test results on unseen encoders with SAR@FAR$(=1e-4, 1e-3, 1e-2)$ and SAR$(score {\geq} threshold)$ for each of the datasets respectively. Best results are highlighted in bold. 
    }
    \centering
    \renewcommand{\arraystretch}{1.2}
    \resizebox{0.7\linewidth}{!}{
    \begin{tabular}{lcccccccccccccc}
    \toprule
          \multicolumn{3}{c}{Dataset} & \multicolumn{4}{c}{\textbf{LFW}} & \multicolumn{4}{c}{\textbf{CFP-FP}} & \multicolumn{4}{c}{\textbf{AgeDB-30}} \\ 
        \midrule
         \multicolumn{3}{c}{Metric }& \multicolumn{3}{c}{\textbf{\emph{SAR@FAR}}}  & \textbf{\emph{SAR}} & \multicolumn{3}{c}{\textbf{\emph{SAR@FAR}}} & \textbf{\emph{SAR}}  & \multicolumn{3}{c}{\textbf{\emph{SAR@FAR}}} & \textbf{\emph{SAR}} \\
        \midrule
        \texttt{\#} Latents & unsup. val. & latent avg.  & 1e-4 & 1e-3 & 1-e2 &  & 1e-4 & 1e-3 & 1-e2 & & 1e-4 & 1e-3 & 1-e2 & \\ 
        \midrule
        \multicolumn{3}{l}{\scshape{Real Face}} & 97.03 & 99.82 & 99.95 & 99.93 & 60.26 & 75.13 & 87.05 & 90.64 & 69.95 & 76.88 & 87.42 & 95.66 \\
        \midrule
        \scshape{1} & \redxmark & \redxmark & 27.92 & 42.86 & 58.5 & 48.66 & 16.63 & 31.24 & 49.04 & 57.94 & 14.28 & 21.39 & 40.9 & 52.52 \\  
        \midrule
        \multirow{4}{*}{\scshape{20}} & \redxmark & \redxmark & 66.64 & 87.78 & 95.28 & 91.61 & 31.86 & 51.91 & 72.74 & 79.97 & 26.94 & 37.06 & 59.82 & 72.9 \\ 
        & \redxmark & \greencheck & 70.06 & 89.91 & 96.18 & 93.18 & 33.48 & 53.33 & 73.45 & 81.26 & 28.21 & 38.6 & 61.91 & 74.62 \\ 
        & \greencheck & \redxmark & 68.42 & 89.89 & 96.39 & 93.24 & 32.91 & 53.15 & 73.25 & 80.95 & 27.74 & 38.47 & 61.56 & 74.66 \\ 
        \rowcolor{Gray} \cellcolor{white} & \greencheck \cellcolor{white}  & \greencheck \cellcolor{white} &  \textbf{69.71} & \textbf{91.48} & \textbf{96.97} & \textbf{94.48} & \textbf{33.64} & \textbf{54.74} & \textbf{74.64} & \textbf{82.01} & \textbf{29.16} & \textbf{39.95} & \textbf{63.2} & \textbf{76.07} \\ 
        \midrule
        \multirow{4}{*}{\scshape{50}} & \redxmark & \redxmark & 67.48 & 88.17 & 95.36 & 91.92 & 32.29 & 52.49 & 72.27 & 80.16 & 27.46 & 37.49 & 60.21 & 73.06 \\ 
        ~ & \redxmark & \greencheck & 70.88 & 90.41 & 96.15 & 93.53 & 33.9 & 53.67 & 74.46 & 82.04 & 28.54 & 38.93 & 61.71 & 74.47 \\ 
        ~ & \greencheck & \redxmark & 70.63 & 91.11 & 96.95 & 94.28 & 34.22 & 54.22 & 74.79 & 81.62 & 28.72 & 39.08 & 62.03 & 75.08 \\ 
        \rowcolor{Gray} \cellcolor{white} & \greencheck \cellcolor{white} & \greencheck \cellcolor{white} & \textbf{71.69} & \textbf{91.64} & \textbf{97.12} & \textbf{94.8} & \textbf{35.02} & \textbf{55.07} & \textbf{75.03} & \textbf{82.46} & \textbf{29.77} & \textbf{40.52} & \textbf{63.65} & \textbf{76.4} \\ 
        \midrule
        \multirow{4}{*}{\scshape{100}} & \redxmark & \redxmark & 68.83 & 88.83 & 95.52 & 92.06 & 33.02 & 53.06 & 72.74 & 80.81 & 28.48 & 38.49 & 61.04 & 73.9 \\ 
        ~ & \redxmark & \greencheck & 70.87 & 90.5 & 96.13 & 93.46 & 34.6 & 55.26 & 74.6 & 81.99 & 29.14 & 39.54 & 62.33 & 74.92 \\ 
        ~ & \greencheck & \redxmark &  72.22 & 91.45 & 96.83 & 94.24 & 35.28 & 55.69 & 75.15 & 82.69 & 29.87 & 40.49 & 63.31 & 75.9  \\ 
        \rowcolor{Gray} \cellcolor{white} & \greencheck \cellcolor{white}& \greencheck \cellcolor{white}&  \textbf{73.04} &  \textbf{92.13} & \textbf{97.36} & \textbf{95.13} & \textbf{36.85} & \textbf{57.25} & \textbf{76.29} & \textbf{82.96} & \textbf{30.86} & \textbf{41.64} & \textbf{64.38} & \textbf{76.98} \\
        \bottomrule
    \end{tabular}
    }
\label{table:ablation_component}
\end{table*}

\subsection{Analysis}
\subsubsection{Ablation of Components}
We analyze the effect of each component of our method. In Tab.~\ref{table:ablation_component}, we present results for $n=1, 20, 50, 100$ with and without latent averaging and unsupervised validation. We use $k_{top}=10$ and $t_{avg}=70$ for default when applied. We additionally report SAR@FAR(=1e-4, 1e-3, 1-e2),
which is the success attack rate at thresholds of FAR.
SAR@FAR is similar to TAR@FAR widely used in T/F evaluations, facilitating a more thorough analysis. Results in Tab.~\ref{table:ablation_component} reveal that using more initial samples and applying ALSUV yield the best performance. Comparing with optimizing a single latent without ALSUV, our method increases SAR by 46.14\% for LFW, 12.23\% for CFP-FP, and 22.62\% for AgeDB-30 on average.

    \subsubsection{Analysis of Hyperparameters}
    We investigate the effects of each hyperparameter: number of latents $n$, size of latent average $t_{avg}$, and number of top $k$ samples for unsupervised validation $k_{top}$. We evaluate results by controlling these hyperparameters tested on unseen encoders on LFW dataset.

\begin{figure*}[t]
\centering
\includegraphics[width=.85\linewidth]{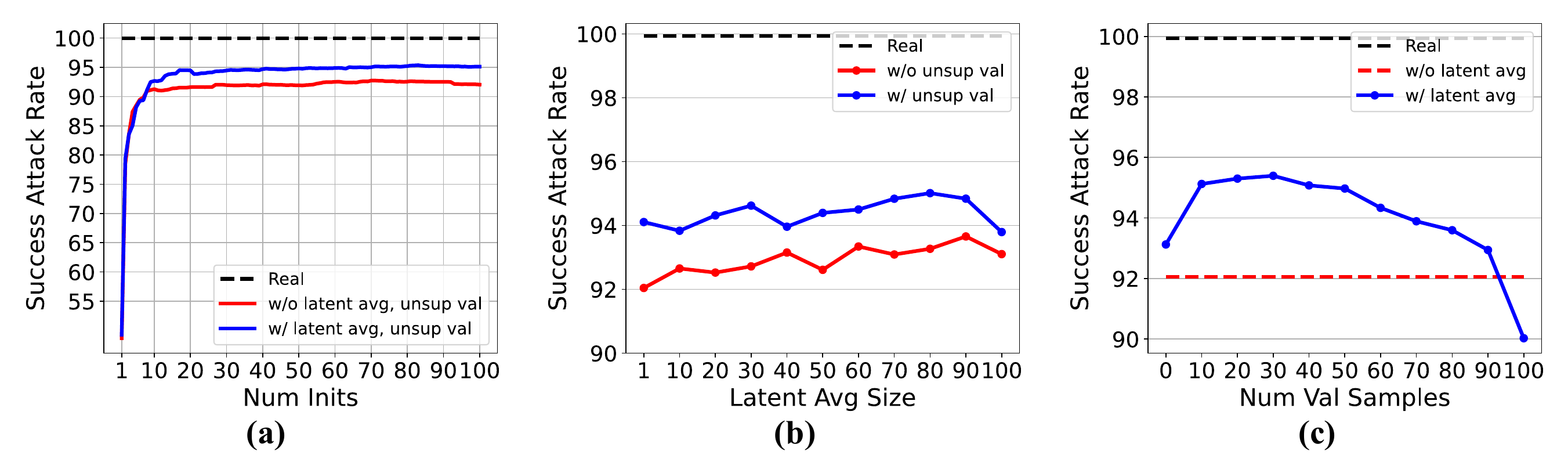}
\caption{
Results of variating hyperparameters (a) number of latents $n$, (b) size of latent averaging $t$, and (c) number of samples $k_{top}$ for unsupervised validation with LFW dataset.
}
\label{fig:ablation_graphs}
\end{figure*}

    First of all, we consider $1 \leq n \leq 100$ with (blue) and without (red) ALSUV. As shown in Fig.~\ref{fig:ablation_graphs}a, $n$ plays a crucial role as SAR increases from 48.66\% to 92.06\% (red line). SAR increases remarkably, especially within the range of $1\leq n\leq 10$, and starts plateauing from $20\leq n$ as shown in Fig.~\ref{fig:ablation_graphs}a and Tab.~\ref{table:ablation_component}. Therefore, we suggest $n=20$ as a cost-efficient trade-off point between high performance and computation. Additionally, the gap between red and blue lines shows the influence of latent averaging and unsupervised validation.

    Next, we analyze the effect of the size of latent average $t_{avg}$ for $t_{avg}\in$ \{1, 10, 20, 30, 40, 50, 60, 70, 80, 90, 100\} with $n=100$ and test with and without unsupervised validation. As shown in Fig.~\ref{fig:ablation_graphs}b, the overall performance is highest for $70\leq t_{avg}\leq90$. Compared with $t_{avg}=1$ (without latent averaging), latent averaging improves the performance from 92.06\% to 93.46\% without unsupervised validation, and 94.24\% to 95.13\% with unsupervised validation. Interestingly, any size of latent averaging benefits without unsupervised validation(red line).

    Finally, we investigate the effect of the number of top $k$ samples $k_{top}$ for unsupervised validation. With $n=100$, we vary $k_{top} \in$ \{1, 10, 20, 30, 40, 50, 60, 70, 80, 90, 100\} with latent average $t=70$(blue line) and without latent average(red dashed line). Even without latent averaging, unsupervised validation increases performance from 92.06\% to 94.24\%. Applying latent averaging increases even more to 95.13\%. As shown in Fig.~\ref{fig:ablation_graphs}c, the best range of $k_{top}$ is $10 \leq k_{top} \leq 30$ where the optimal value is $k_{top}=30$ and practical value is $k_{top}=10$. Overall, we suggest $k_{top}$ as 10\% to 30\% of the number of latents $n$. 

\begin{figure}[t]
\centering
\includegraphics[width=.75\linewidth]{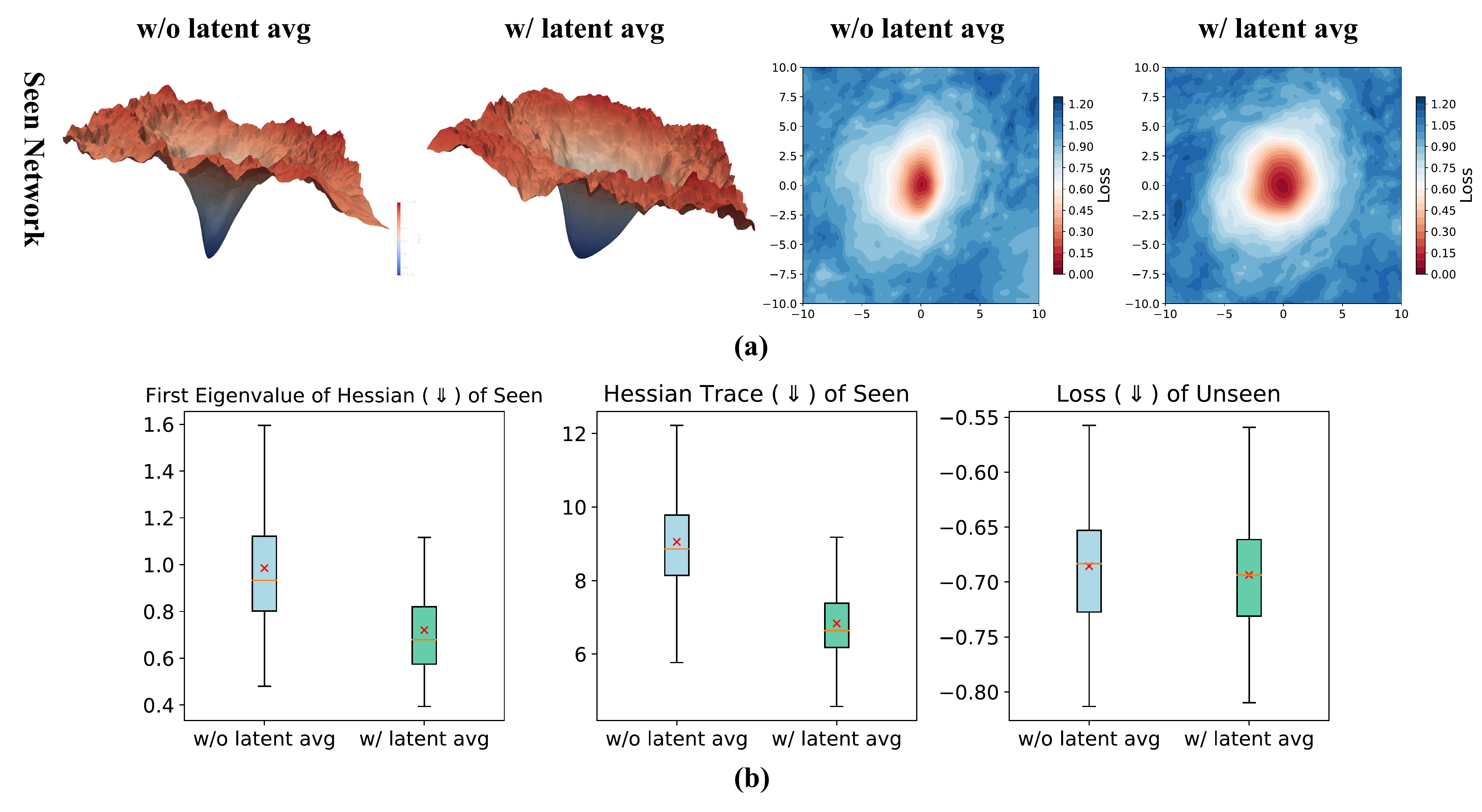}
\caption{Loss surface landscape and contour line as heat map(a) of a single sample seen encoder with and without latent averaging applied. (b) Quantitative statistics are shown as box plot showing the first eigenvalue of hessian(left), trace of hessian(middle) for seen encoder and loss for unseen encoders measured with LFW samples.
}
\label{fig:loss_surf}
\end{figure}

\begin{table}[t]
\centering
\begin{minipage}[t]{0.395\linewidth}
\caption{
Averaged SAR of unseen encoders on LFW dataset. We compare multiple (denoted as 100$\times$100) with single (denoted as 1$\times$10k) latent optimization with periodic learning rate scheduler and the corresponding loss.
}
\label{table:ablation_periodic}
\centering
    \resizebox{0.95\linewidth}{!}{
    \begin{tabular}{ll|ccc|c}
    \toprule
         \multirow{2}{*}{Target Encoder}&  Method  & \multicolumn{3}{c|}{\textbf{\emph{SAR@FAR}}}  & \multirow{2}{*}{\textbf{\emph{SAR}}}   \\ 
         & ($n \times iter$) & 1e-4 & 1e-3 & 1e-2 &  \\ 
         \midrule
         & $ 1 \times 10k$ & 9.44 & 22.15 & 30.08 & 26.22 \\
         \rowcolor{Gray} \cellcolor{white}\multirow{-2}{*}{FaceNet}& $100\times100$ & 61.79 & 79.75 & 91.2 & 86.3 \\ 
         \midrule
         & $1 \times 10k$ & 59.01 & 82.72 & 89.76 & 85.92 \\
         \rowcolor{Gray} \cellcolor{white}\multirow{-2}{*}{MobFaceNet} & $100\times100$ & 69.1 & 94.5 & 99.01 & 97.62 \\ 
         \midrule
         & $1 \times 10k$ & 22.8 & 29.7 & 34.87 & 32.29 \\
         \rowcolor{Gray} \cellcolor{white}\multirow{-2}{*}{ResNet50}& $100\times100$ & 75.76 & 92.69 & 97.51 & 95.81 \\ 
         \midrule
         & $1 \times 10k$ &  55.56 & 72.68 & 78.99 & 76.13 \\
         \rowcolor{Gray} \cellcolor{white}\multirow{-2}{*}{ResNet100} & $100\times100$ & 80.35 & 97 & 99.31 & 98.91 \\
         \midrule
         & $ 1 \times 10k$ & 49.31 & 66.36 & 71.32 & 69.48 \\ 
         \rowcolor{Gray} \cellcolor{white}\multirow{-2}{*}{Swin-S} & $100\times100$& 76.07 & 97.32 & 99.43 & 99.07 \\
         \midrule
         & $ 1 \times 10k$ & 54.86 & 68.34 & 75.52 & 71.62 \\
         \rowcolor{Gray} \cellcolor{white}\multirow{-2}{*}{VGGNet} & $100\times100$ & 75.15 & 89.14 & 96.37 & 92.94 \\
         \midrule
         \midrule
         Avg Cos Sim & $ 1 \times 10k$ & \multicolumn{4}{c}{0.8842} \\
         \rowcolor{Gray} \cellcolor{white}at seen encoder & $100\times100$ & \multicolumn{4}{c}{0.9813} \\
        \bottomrule
    \end{tabular}
    }
\end{minipage}
\hfill
\begin{minipage}[t]{0.595\linewidth}
\caption{
Comparing result with DiBiGAN\cite{duong2020vec2face} on each datasets. Experiments were done under the same white-box setting. We used 20 parallel latents to compare with \cite{duong2020vec2face}.
}
\label{table:compare_vec2face}
\centering
\resizebox{0.65\linewidth}{!}{
    \begin{tabular}{llccc}
    \toprule
        \multirow{2}{*}{Target Encoder} & \multirow{2}{*}{Method} & \multicolumn{3}{c}{Dataset}  \\ 
    \cmidrule{3-5}
         &  & LFW & CFP-FP & AgeDB-30 \\ 
    \midrule
        \multirow{3}{*}{ResNet100} & \scshape{Real Face} & 99.96 & 97.1 & 98.4 \\ 
         & \scshape{DiBiGAN} & 99.18 & 92.67 & 94.18 \\ 
         \rowcolor{Gray}\cellcolor{white} & \scshape{Ours} & 99.85 & 96.59 & 98.33 \\
    \bottomrule
    \end{tabular}
}
\bigskip
\caption{
Quantitative face quality evaluation using 2 different types of FIQAs where higher values are better. Best results are highlighted with bold and second are underlined.
}
\label{table:face_quality}
\centering
\resizebox{0.995\linewidth}{!}{
    \begin{tabular}{lccccccccc}
    \toprule
        Metric & ~~Real~~ & NBNet & LatentMap & Genetic & GaussBlob & EigenFace & FaceTI & QEZOGE & Ours  \\ 
    \midrule
        SER-FIQ($\uparrow$) & 0.774 & 0.612 & 0.661 & 0.664 & 0.005 & 0.519 & \textbf{0.748} & 0.701 & \underline{0.746} \\ 
        CR-FIQA($\uparrow$) & 2.191 & 1.756 & 2.139 & \underline{2.181} & 0.747 & 1.723 & 2.198 & \textbf{2.262} & 2.107 \\
    \bottomrule
    \end{tabular}
}
\end{minipage}
\end{table}

    \subsubsection{Number of Latents and Optimization Steps}
    We compare optimizing 100 latents optimized 100 steps each (denoted as 100 $\times$ 100, without latent average and unsupervised validation) and a single latent optimized 10,000 steps with cyclical learning rate\cite{izmailov2018averaging}. We train on LFW dataset where results are shown in Tab.~\ref{table:ablation_periodic}. Our method outperforms with 95.13\% in average SAR while serial optimization shows 60.28\% resulting in a 34.85\% performance gap. Despite the total steps of optimization being identical, results highlight the importance of using multiple latents which prevents falling into poor local minima than interacting with complex learning rate scheduler as the difference of average cosine similarity shown in the last row of Tab.~\ref{table:ablation_periodic} is significant.
    
    \subsubsection{Latent Averaging and Loss Surface}
    In this section, we thoroughly analyze the effect of latent averaging. First of all, we visualize the loss surface of a sample in the seen encoder by adding perturbation in two random axes and using the method suggested in \cite{li2018visualizing}. Additionally, we quantitatively compare the flatness of the optima point by measuring the first eigenvalue and trace of the Hessian matrix for every sample in LFW and the generalization by loss value on unseen encoders. Fig.~\ref{fig:loss_surf}a shows the shape of the loss surface where the z-axis is the loss value $1-sim(\cdot)$ where 1 is added to set the minimum loss value to 0. Applying latent average significantly improves the flatness visually as shown. In addition, the statistics of the first eigenvalue and trace of the Hessian matrix of latents with latent averaging have much lower values which indicate the curvature at the minima point is flatter while the loss value for unseen encoders is lower.

    \subsubsection{Unsupervised Validation with Pseudo Target}
\begin{figure*}[t]
\centering
\includegraphics[width=.875\linewidth]{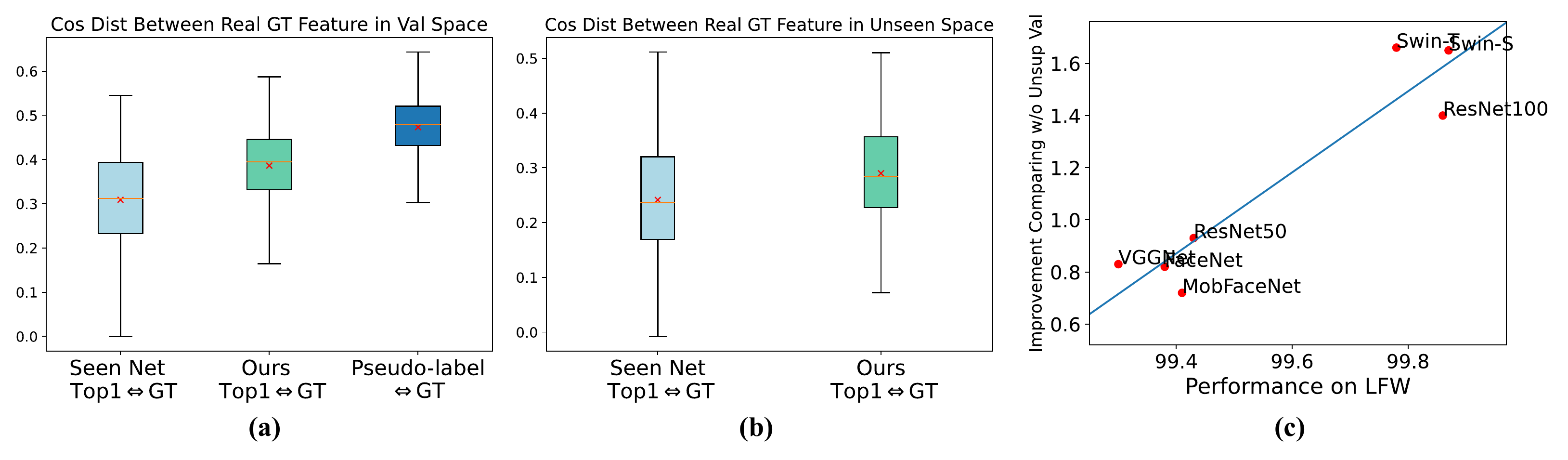}
\caption{Cosine distance statistics between seen encoder's top 1, unsupervised validation's top 1, and pseudo target against real image's feature in validation feature space (a) and unseen encoder space (b) (pseudo target excluded since it only presents in validation space). (c) shows the correlation between the validation encoder's performance and performance improvement.
88}
\label{fig:ulv_ablation}
\end{figure*}

    Unsupervised validation utilizes the feature space of a surrogate validation encoder to search for a better generalizing sample instead of only searching the seen encoder feature space. Hence, we compare the distance between the validation encoder's top 1 and seen encoder's top 1 against the ground truth feature in validation space (Fig.~\ref{fig:ulv_ablation}a) and examine whether this is relevant to generalizing to unseen encoder space (Fig.~\ref{fig:ulv_ablation}b). We also measure the distance between pseudo targets against ground truth in validation space to verify its efficacy as a target. We use LFW dataset and examine all 6 encoders as seen and unseen between each other as Tab.~\ref{table:result_sar} with Swin-T as validation encoder. 
    
    As shown in Fig.~\ref{fig:ulv_ablation}a the pseudo target is closest to the ground truth feature from real images with average cosine distance $0.473$ followed by our unsupervised validation's top 1 with $0.386$ and seen encoder's top 1 with $0.309$ in the validation space. The closest pseudo target might be the best option to use, but unfortunately, its corresponding latent is inaccessible which makes the sample closest to the pseudo target our best choice. This aspect is connected to the unseen encoder space in Fig.~\ref{fig:ulv_ablation}b where our method shows average cosine distance $0.29$ higher than seen encoder's top 1 $0.241$. 
    
    Furthermore, we examine performances by varying the type of validation encoder. Each encoders used in our work are used as validation encoders and performance improvement only for unseen cases where seen, unseen, and validation encoders do not overlap are measured. Results in Fig.~\ref{fig:ulv_ablation}c presents that using any validation encoder shows improvement with positive correlation with the general performance of the encoder. More results are shown in Supp.~\ref{sec:supp_results} Tab.~\ref{table:result_ulv}.

\begin{figure*}[t]
\centering
\includegraphics[width=.9\linewidth]{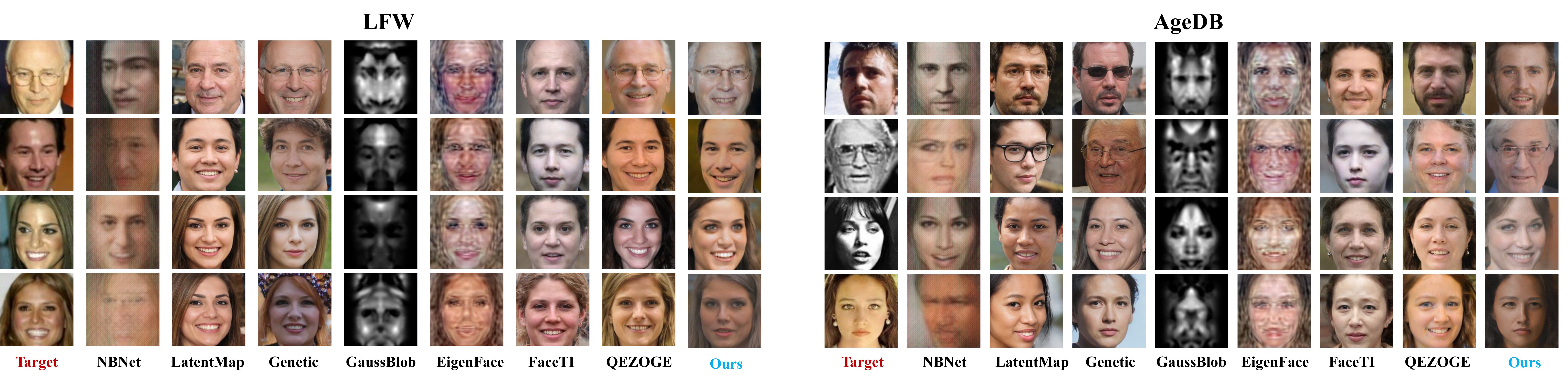}
\caption{
Sample images from baseline methods and ours on LFW and AgeDB. 
}
\label{fig:sample_images}
\end{figure*}

\subsubsection{Image Quality}
We conduct qualitative and quantitative analyses of generated images. For quality evaluation, we compare a few samples from LFW and AgeDB in Fig.~\ref{fig:sample_images} and CFP-FP in Supp.~\ref{sec:supp_results} Fig.~\ref{fig:sample_cfp}. For quantitative evaluation, we adopt face-specific quality metrics SER-FIQ \cite{terhorst2020ser} and CR-FIQA \cite{boutros2023cr}. NBNet, GaussBlob, and EigenFace show poor image quality visually with artifacts and quantitatively low image quality metrics. Meanwhile, despite the decent quality, LatentMap, Genetic, and FaceTI present wrong identities compared with the target images. On the other hand, QEZOGE and our method show decent image quality visually, quantitatively, and content-wise as shown in Tab.~\ref{table:face_quality}.

\section{Conclusion}
In this paper, we have presented a framework for face reconstruction transfer attacks. We devised our method ALSUV inspired by out-of-distribution generalization to generalize our generated sample to unseen face encoders. ALSUV is instantiated by combining multiple latent optimization, latent averaging, and unsupervised validation with the pseudo target. We demonstrate that our approach surpasses previous methods in FRTA by showing high SAR and identification rate across various unseen face encoders. Our thorough analysis shows the effectiveness of our method inspired by OOD generalization. Furthermore, we hope our work alerts the security risk posed by FRTA, and emphasizes the awareness to mitigate potential threats.

\sloppy
\section*{Acknowledgements}
This work is supported in part by NSF grant CNS-2038493, ONR grant N00014-21-1-2431 from NCI, and U.S. Department of Homeland Security grant 22STESE00001-03-02. The views and conclusions contained in this document are those of the authors and should not be interpreted as necessarily representing the official policies, either expressed or implied, of the U.S. Department of Homeland Security, the NSF and the ONR.

\bibliographystyle{splncs04}
\bibliography{main}

\clearpage

\appendix
\onecolumn

\section{Proof to Theorem}
\label{sec:supp_proof}

\noindent
\textbf{Theorem.}
Define $f_z$ by $f_z(\theta) = E_\theta( G(z))$,
and let $D^*_{seen}=\{ (\theta_{seen}, v_{\theta_{seen}}) \}$, $\mathcal{D}^* = \{ \{ (\theta, v_{\theta}) \} : \theta \in \Theta \}$, and
\begin{equation}
l(f_z(\theta), v_\theta) = - sim(f_z(\theta), v_\theta).
\end{equation}
Then, $f_z$ is an MLP, and the FRTA algorithm $\mathcal{A}$ on $\Theta$ is an OOD generalization algorithm $\mathcal{A}^*$ on the domain $\mathcal{D}^*$ in the parameter space $\mathcal{Z}$.

\begin{proof}
(MLP)
Since MLP is a composition of MLPs, it suffices to prove that a layer of $f_z$ is MLP. To this end, we show that the MLP $\sigma ( \mathbf{W}\mathbf{x} + \mathbf{b})$
with input $\mathbf{x}$ and parameters $(\mathbf{W}, \mathbf{b})$ is an MLP with input $(\mathbf{W}, \mathbf{b})$ and parameters $\mathbf{x}$. 
Let $\mathbf{w}_i$  and $b_i$ denote the $i$-th row of $\mathbf{W}$ and $\mathbf{b}$, respectively, for $i=1,\dots, r$ where $r$ is the row dimension of $\mathbf{W}$. Observe,
\begin{equation}
\sigma ( \mathbf{W}\mathbf{x} + \mathbf{b})
=
\sigma
\left(
diag(\mathbf{x}) \mathbf{w}_i  + b_i \mathbf{1}
\right)
\end{equation}
where $diag(\mathbf{x})$ is the diagonal matrix whose diagonal elements are $x_i$, and $\mathbf{1}$ is a vector whose all entries are $1$. Both
\begin{equation}
f_1((\mathbf{W}, \mathbf{b}); \mathbf{x}) = diag(\mathbf{x}) \mathbf{w}_i    
\end{equation}
and 
\begin{equation}
f_2((\mathbf{W}, \mathbf{b}); \mathbf{x}) = b_i \mathbf{1}
\end{equation}
are MLPs with input $(\mathbf{W}, \mathbf{b})$ and parameters $\mathbf{x}$, hence their sum and activattion are also MLPs with the same aspect, completing the proof.

(Equivalence)
We show the equivalence between FRTA and OOD generalization. To see this, first define
\begin{equation}
L(z; D^*) := \frac{1}{|D^*|}
\sum_{(\theta, v_\theta) \in D^*}
l(f_z(\theta), v_\theta)
\end{equation}
Then,
observe that
\begin{align*}
\mathcal{A}^* (D^*_{seen}) &:= 
\min_z \max_{D^* \in \mathcal{D}^*} L(z; D^*) \\
&= \min_z \max_{D^* \in \mathcal{D}^*}
l(f_z(\theta), v_\theta) \\
&= \min_z \max_{D^* \in \mathcal{D}^*} - sim (f_z(\theta), v_\theta) \\
&= \max_z \min_{\theta \in \Theta} sim (f_z(\theta), v_{\theta}) \\
&=: \mathcal{A}(\theta_{seen})
\end{align*}
where the second and fourth equations hold due to $D^* = \{ (\theta, v_\theta) \}$, completing the proof.
\end{proof}

\section{Supplementary to Method}
\label{sec:supp_method}

\subsection{Algorithm of the full method}
The full algorithm of our method is given in Algorithm.~\ref{alg:method}. In this algorithm, $[z_i]_{i=1}^n$ is the vector concatenation of the vectors $z_i$, which is to parallelize the update of $z_i$'s.

\begin{algorithm}
\caption{The algorithm of our method}
\label{alg:method}
\begin{algorithmic}[1]
\Require $\{z_i\}_{i=1}^n, E_{\theta_{seen}}, G, v_{seen}, T, T_0, E_{\theta_{val}}, k_{top}$
\Ensure $z^*$
\Statex \# Optimizing Multiple Latents
\State Initialize $z_i^{(0)}= z_i$  for $i=1,\dots,n$
\For{$t=1,\dots, T$}
    \State $[z_i^{(t)}]_{i=1}^n \leftarrow [z_i^{(t-1)}]_{i=1}^n + \lambda \nabla_{[z_i]_{i=1}^n} \sum_{i=1}^n \simop (E_{\theta_{seen}}(G(z_i)), v_{seen}) \mid_{[z_i]_{i=1}^n = [z_i^{(t-1)}]_{i=1}^n}$
    \Comment{by Adam optimizer}
\EndFor
\Statex \# Latent Averaging 
\State $\overline{z}_{i} \leftarrow \frac{1}{T_0} \sum_{t = T - T_0}^{T} z_i^{(t)}$
\Statex \# Unsupervised Validation with Pseudo Target
\State Order the index $i$ of $\overline{z}_{i}$ such that 
$\simop(E_{\theta_{seen}}(G(\overline{z}_{i+1})), v_{seen}) \leq \simop(E_{\theta_{seen}}(G(\overline{z}_{i})), v_{seen})$
\State $ m = \frac{1}{k_{top}} \sum_{i=1}^{k_{top}} E_{\theta_{val}}(G(\overline{z}_{i}))$
\Comment{pseudo target to the validation encoder embedding of real image}
\State 
 $z^* =  \underset{z \in \{ \overline{z}_{i} \}_{i=1}^{k_{top}} }{\arg \max} \simop (E_{\theta_{val}}(G(z)), m)$
\end{algorithmic}
\end{algorithm}

\clearpage

\section{Supplementary Results}
\subsection{Sample Images}
\label{sec:supp_results}
\begin{figure*}
\centering
\includegraphics[width=.7\linewidth]{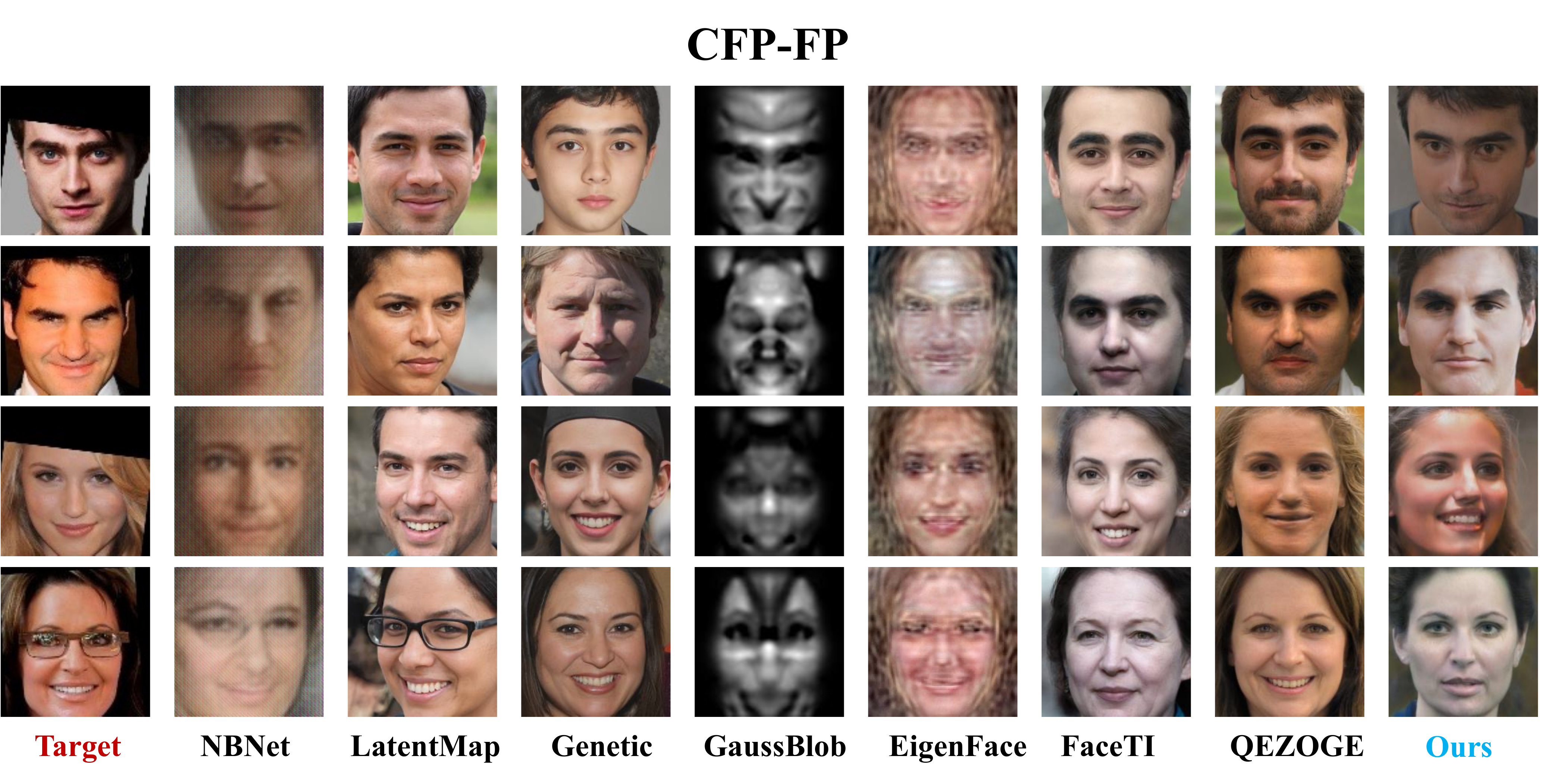}
\caption{
Sample images of previous methods and ours from CFP-FP dataset. 
}
\label{fig:sample_cfp}
\end{figure*}

Fig.~\ref{fig:sample_cfp} shows the result of reconstructed images from CFP-FP dataset of each baselines, our method and ground truth images. 

\subsection{Unsupervised Validation Analysis}

Tab.~\ref{table:result_ulv} ablates pseudo target of unsupervised validation by using different types of target for searching the top 1 reconstructed sample. We compare 3 different cases: using the seen feature as target in the seen encoder space, using validation encoder and the pseudo target in the validation encoder space, and using the unseen encoder and the feature from real image in the unseen encoder space where the last works as a reference to upper bound performance.

\begin{table*}
    \caption{
    SAR comparison by using different targets when choosing top 1 sample. Unsupervised validation results are highlighted with gray and best performances (except upper bound) are highlighted with bold. 
    }
    \centering
    \resizebox{0.7\linewidth}{!}{
    \begin{tabular}{llccccccc}
    \toprule 
         \multicolumn{2}{c}{Dataset}  &  \multicolumn{7}{c}{\textbf{LFW}}   \\
        \midrule
        Target Encoder& Method & FaceNet & MobFace & ResNet50 &ResNet100 & Swin-S & VGGNet & Unseen AVG \\ 
        \midrule
         & Seen Feature & \textbf{99.87}\cellcolor{red!25} & 55.81 & 97.44 & 81.53 & 79.84 & \textbf{99.39} & 82.8 \\ 
         \rowcolor{Gray} \cellcolor{white}& Pseudo Target & 99.83\cellcolor{red!25} & \textbf{65.15} & \textbf{97.51} & \textbf{86.79} & \textbf{83.32} & 99.22 & \textbf{86.4} \\
         \multirow{-3}{*}{FaceNet}& Upper bound & 99.87\cellcolor{red!25} & 91.51 & 99.19 & 97.34 & 95.25 & 99.66 & 96.59 \\ 
        \midrule
         & Seen Feature & 95.72 & \textbf{99.87}\cellcolor{red!25} & 96.6 & 96.73 & 99.09 & 97.88 & 97.2 \\ 
         \rowcolor{Gray} \cellcolor{white}& Pseudo Target & \textbf{96.76} & 99.83\cellcolor{red!25} & \textbf{97.3} & \textbf{96.83} & \textbf{99.16} & \textbf{98.05} & \textbf{97.62}  \\ 
         \multirow{-3}{*}{MobFace}& Upper bound & 98.75 & 99.87\cellcolor{red!25} & 99.26 & 99.53 & 99.6 & 99.19 & 99.27   \\ 
        \midrule
         & Seen Feature & 98.99 & 77.15 & 99.76\cellcolor{red!25} & 96.49 & 95.38 & \textbf{99.73} & 93.55  \\ 
         \rowcolor{Gray} \cellcolor{white}& Pseudo Target & \textbf{99.12} & \textbf{85.24} & \textbf{99.8}\cellcolor{red!25} & \textbf{97.44} & \textbf{97.61} & 99.63 & \textbf{95.81}  \\ 
         \multirow{-3}{*}{ResNet50}& Upper bound & 99.49 & 95.96 & 99.76\cellcolor{red!25} & 99.63 & 99.36 & 99.63 & 98.81   \\ 
        \midrule
         & Seen Feature & \textbf{99.39} & 93.93 & \textbf{99.6} & 99.93\cellcolor{red!25} & 99.87 & 99.53 & 98.46  \\
         \rowcolor{Gray} \cellcolor{white}& Pseudo Target & 99.36 & \textbf{96.16} & 99.56 & \textbf{99.93}\cellcolor{red!25} & \textbf{99.93} & \textbf{99.56} & \textbf{98.91} \\ 
         \multirow{-3}{*}{ResNet100}& Upper bound & 99.53 & 99.36 & 99.8 & 99.93\cellcolor{red!25} & 99.93 & 99.7 & 99.66  \\ 
        \midrule
         & Seen Feature & \textbf{98.42} & 98.18 & \textbf{99.12} & 99.73 & 99.93\cellcolor{red!25} & 99.33 & 98.96  \\ 
        \rowcolor{Gray} \cellcolor{white}& Pseudo Target & 98.08 & \textbf{99.02} & 99.02 & \textbf{99.87} & \textbf{99.93}\cellcolor{red!25} & \textbf{99.36} & \textbf{99.07} \\ 
         \multirow{-3}{*}{Swin-S}& Upper bound & 99.36 & 99.56 & 99.56 & 99.93 & 99.93\cellcolor{red!25} & 99.66 & 99.61  \\ 
        \midrule
         & Seen Feature & 99.26 & 66.33 & 99.49 & 93.46 & 89.15 & 99.76\cellcolor{red!25} & 89.54   \\ 
         \rowcolor{Gray} \cellcolor{white}& Pseudo Target & \textbf{99.29} & \textbf{77.45} & \textbf{99.49} & \textbf{94.57} & \textbf{93.9} & \textbf{99.8}\cellcolor{red!25} & \textbf{92.94}  \\ 
         \multirow{-3}{*}{VGGNet}& Upper bound & 99.6 & 97.17 & 99.73 & 99.09 & 98.85 & 99.8\cellcolor{red!25} & 98.89 \\ 
    \bottomrule
    \end{tabular}
    }
\label{table:result_ulv}
\end{table*}

\clearpage

\subsection{Attacking SOTA Face Recognition Systems}

\begin{table*}[t]
    \caption{SAR of previous works and our proposed method with SOTA target encoders: AdaFace \cite{Kim_2022_CVPR}, CurricularFace \cite{Huang_2020_CVPR}, and ElasticFace \cite{Boutros_2022_CVPR}. We highlight our method with grey and best results are highlighted with bold.}
    \centering
    \resizebox{0.97\linewidth}{!}{
    \begin{tabular}{ll|cccccc|c}
    \toprule 
         \multicolumn{2}{c|}{Dataset} &  
         \multicolumn{7}{c}{\textbf{LFW}}  \\
        \midrule
         \multicolumn{2}{c|}{Test Encoder}  & 
         FaceNet & MobFace & ResNet50 & ResNet100 & Swin-S & VGGNet & Unseen AVG \\ 
        \midrule
        Target Encoder & \scshape{Real Face} & 
        99.87 & 99.83 & 99.83 & 99.93 & 99.93 & 99.8 & - \\ 
        \midrule

        \multirow{8}{*}{AdaFace} 
         & \scshape{NBNet} & 
         49.27 & 12.99 & 61.56 & 20.18 & 36.31 & 41.92 & 37.04 \\
         & \scshape{LatentMap} & 
         15.36 & 2.06 & 13.4 & 3.88 & 1.55 & 9.15 & 7.57 \\ 
         & \scshape{Genetic} & 
         14.17 & 3.58 & 7.19 & 7.49 & 12.22 & 8.23 & 8.81 \\
         & \scshape{GaussBlob} & 
         0.03 & 30.14 & 0.07 & 85.59 & 10.29 & 0.03 & 21.03 \\
         & \scshape{EigenFace} & 
         32.97 & 80.16 & 68.51 & 98.31 & 66.86 & 77.49 & 70.72 \\
         & \scshape{FaceTI} & 
         0.67 & 7.93 & 3.24 & 26.49 & 52.55 & 0.54 & 15.24 \\
         & \scshape{QEZOGE} & 
         73.41 & 63.75 & 72.7 & 83.8 & 83.53 & 74.65 & 75.31 \\
         \rowcolor{Gray} \cellcolor{white} & \scshape{Ours} &  
         \textbf{88.3} & \textbf{96.8} & \textbf{97.37} & \textbf{99.33} & \textbf{99.09} & \textbf{95.75} & \textbf{96.11} \\
        \midrule

        \multirow{8}{*}{CurricularFace} 
         & \scshape{NBNet} & 
         58.72 & 10.53 & 63.99 & 58.18 & 39.79 & 46.81 & 46.34 \\
         & \scshape{LatentMap} & 
         28.99 & 5.26 & 19.14 & 9.52 & 4.22 & 14.38 & 13.59 \\
         & \scshape{Genetic} & 
         31.69 & 8.2 & 17.25 & 21.53 & 23.96 & 19.27 & 20.32 \\
         & \scshape{GaussBlob} & 
         0.64 & 27.98 & 0.51 & 86.6 & 6.65 & 0.37 & 20.46 \\
         & \scshape{EigenFace} & 
         51.54 & 79.95 & 79.24 & 97.91 & 56.06 & 82.52 & 74.54 \\
         & \scshape{FaceTI} & 
         2.36 & 16.06 & 9.58 & 47.65 & 58.42 & 1.92 & 22.67 \\
         & \scshape{QEZOGE} & 
         86.2 & 67.09 & 80.97 & 90.08 & 84.75 & 82.52 & 81.94 \\
         \rowcolor{Gray} \cellcolor{white} & \scshape{Ours} &  
         \textbf{93.97} & \textbf{95.99} & \textbf{96.73} & \textbf{99.29} & \textbf{98.38} & \textbf{95.99} & \textbf{96.73} \\
        \midrule

        \multirow{8}{*}{ElasticFace} 
        & \scshape{NBNet} & 
        50.86 & 6.78 & 65.47 & 57.34 & 39.66 & 45.87 & 44.33 \\
         & \scshape{LatentMap} & 
         27 & 5.13 & 16.94 & 8.98 & 4.32 & 14.14 & 12.75 \\
         & \scshape{Genetic} & 
         30.81 & 7.9 & 14.88 & 23.08 & 19.44 & 15.86 & 18.66 \\
         & \scshape{GaussBlob} & 
         0.03 & 12.42 & 0.34 & 84.81 & 5.16 & 0.17 & 17.16 \\
         & \scshape{EigenFace} & 
         52.28 & 71.25 & 75.13 & 96.69 & 54.88 & 81.3 & 71.92 \\
         & \scshape{FaceTI} & 
         3.1 & 15.52 & 9.75 & 53.39 & 59.4 & 2.26 & 23.90 \\
         & \scshape{QEZOGE} & 
         82.79 & 66.66 & 81.4 & 92.2 & 84.58 & 81.07 & 81.45 \\
         \rowcolor{Gray} \cellcolor{white} & \scshape{Ours} &  
         \textbf{94.24} & \textbf{95.75} & \textbf{97.03} & \textbf{99.29} & \textbf{98.21} & \textbf{96.49} & \textbf{96.84} \\
        
    \bottomrule
    \end{tabular}
    }
\label{table:result_sota_fr_systems}
\end{table*}

Additionally, we validate the efficacy of ALSUV by attacking SOTA face recognition systems: AdaFace \cite{Kim_2022_CVPR}, CurricularFace \cite{Huang_2020_CVPR}, and Elastic \cite{Boutros_2022_CVPR} with the samples generated from the 6 encoders from Tab.~\ref{table:result_sar} and LFW dataset. ALSUV is compared with previous methods and all resuls are unseen scenario as all 6 encoders do not overlap with the additional face recognition systems. Results are shown in Tab.~\ref{table:result_sota_fr_systems} where ALSUV shows the best results.

\clearpage

\subsection{Additional CFP-FP Results}

\begin{table*}[t]
    \caption{Additional SAR results of CFP-FP pairs compared with previous works with 800 pairs in total. We highlight our method with grey and best results are highlighted with bold.}
    \centering
    \resizebox{0.65\linewidth}{!}{
    \begin{tabular}{ll|cccccc|c}
    \toprule 
         \multicolumn{2}{c|}{Dataset} &  
         \multicolumn{7}{c}{\textbf{CFP-FP}}  \\
        \midrule
         \multicolumn{2}{c|}{Test Encoder}  & 
         FaceNet & MobFace & ResNet50 & ResNet100 & Swin-S & VGGNet & Unseen AVG \\ 
        \midrule
        Target Encoder & \scshape{Real Face} & 
        99.87 & 99.83 & 99.83 & 99.93 & 99.93 & 99.8 & - \\ 
        \midrule

        \multirow{8}{*}{FaceNet} 
         & \scshape{NBNet} & 
         82\cellcolor{red!25} & 20.88 & 74.88 & 40.75 & 16.38 & 70.25 & 44.63 \\
         & \scshape{LatentMap} & 
         51.88\cellcolor{red!25} & 19.50 & 36.75 & 10.75 & 26.25 & 43.88 & 27.43 \\
         & \scshape{Genetic} & 
         60 \cellcolor{red!25}& 23.25 & 41.62 & 10.62 & 29.25 & 42.12 & 29.37 \\
         & \scshape{GaussBlob} & 
         4.88\cellcolor{red!25} & 21 & 8 & 1.38 & 3.12 & 2.75 & 7.25 \\
         & \scshape{EigenFace} & 
         85\cellcolor{red!25} & 28.38 & 58.13 & 18.75 & 27 & 58.75 & 38.20 \\
         & \scshape{FaceTI} & 
         9.75\cellcolor{red!25} & 5.62 & 8.25 & 0.62 & 7.5 & 8.62 & 6.12 \\
         & \scshape{QEZOGE} & 
         93.75\cellcolor{red!25} & 39.12 & 81.62 & 62 & 65 & 83.62 & 66.27 \\
         \rowcolor{Gray} \cellcolor{white} & \scshape{Ours} &  
         \textbf{94.5} & \textbf{47} & \textbf{87.38} & \textbf{74.12} & \textbf{77.12} & \textbf{90.12} & \textbf{75.15} \\
        \midrule

        \multirow{8}{*}{MobileFaceNet} 
         & \scshape{NBNet} & 
         11.75 & 52\cellcolor{red!25} & 17.88 & 8 & 12.38 & 11.62 & 12.33 \\
         & \scshape{LatentMap} & 
         13.12 & 16.75\cellcolor{red!25} & 13 & 2.62 & 13.25 & 15 & 11.40 \\
         & \scshape{Genetic} & 
         18 & 35.38\cellcolor{red!25} & 16.5 & 5.38 & 18.12 & 16.88 & 14.98 \\
         & \scshape{GaussBlob} & 
         8.12 & 62.38\cellcolor{red!25} & 12.25 & 5.25 & 11.62 & 10.75 & 9.60 \\
         & \scshape{EigenFace} & 
         52.25 & 78.62\cellcolor{red!25} & 55.5 & 45.38 & 67.12 & 55.38 & 55.13 \\ 
         & \scshape{FaceTI} & 
         23.88 & 26.12\cellcolor{red!25} & 24.12 & 8.88 & 20.75 & 22.5 & 20.03 \\
         & \scshape{QEZOGE} & 
         45.25 & 74.62\cellcolor{red!25} & 50 & 39 & 57.75 & 49.12 & 48.22 \\
         \rowcolor{Gray} \cellcolor{white} & \scshape{Ours} &  
         \textbf{85.25} & \textbf{82.88} & \textbf{87} & \textbf{86.75} & \textbf{87} & \textbf{86.88} & \textbf{86.58} \\
        \midrule

        \multirow{8}{*}{ResNet50} 
        & \scshape{NBNet} & 
        68.25 & 25.5 & 80.12\cellcolor{red!25} & 49.75 & 10.25 & 68.88 & 44.53 \\
         & \scshape{LatentMap} &
         30.75 & 18.5 & 32.88\cellcolor{red!25} & 10.25 & 22.25 & 28.88 & 22.13 \\
         & \scshape{Genetic} & 
         38.62 & 24.25 & 54.37\cellcolor{red!25} & 11.62 & 28.25 & 39.75 & 28.5 \\
         & \scshape{GaussBlob} & 
         1.75 & 18.5 & 6.5\cellcolor{red!25} & 1 & 2.88 & 3.5 & 5.53 \\
         & \scshape{EigenFace} & 
         68.25 & 38.88 & 83.88\cellcolor{red!25} & 46.88 & 49 & 73.12 & 55.23 \\
         & \scshape{FaceTI} & 
         20.88 & 13.25 & 22.12\cellcolor{red!25} & 4.62 & 14.25 & 18.12 & 14.22 \\
         & \scshape{QEZOGE} & 
         79.12 & 42.25 & 89.12\cellcolor{red!25} & 61.88 & 67.5 & 80.38 & 66.23 \\
         \rowcolor{Gray} \cellcolor{white} & \scshape{Ours} &  
        \textbf{92.62} & \textbf{66} & \textbf{94.38} & \textbf{85.75} & \textbf{86.62} & \textbf{90.88} & \textbf{84.37} \\
        \midrule    

        \multirow{8}{*}{ResNet100} 
         & \scshape{NBNet} & 
         33.75 & 27.88 & 46.38 & 56\cellcolor{red!25} & 27.5 & 40.5 & 35.20 \\
         & \scshape{LatentMap} & 
         21.38 & 14.88 & 18.25 & 10.5\cellcolor{red!25} & 22.88 & 20.38 & 19.55 \\
         & \scshape{Genetic} & 
         27 & 21 & 22.88 & 24.5\cellcolor{red!25} & 28.12 & 25 & 24.8 \\
         & \scshape{GaussBlob} & 
         33.88 & 35.25 & 45.25 & 60.75\cellcolor{red!25} & 47.38 & 37.75 & 39.90 \\
         & \scshape{EigenFace} & 
         72 & 56 & 72.88 & 88.25\cellcolor{red!25} & 81 & 71.62 & 70.7 \\
         & \scshape{FaceTI} & 
         31.87 & 42.88 & 41.62 & 52.12\cellcolor{red!25} & 57.88 & 39.62 & 42.77 \\
         & \scshape{QEZOGE} & 
         56.62 & 41.5 & 60.38 & 88\cellcolor{red!25} & 69.5 & 61.5 & 57.9 \\ 
         \rowcolor{Gray} \cellcolor{white} & \scshape{Ours} &  
         \textbf{92.5} & \textbf{74.25} & \textbf{92.75} & \textbf{93.62} & \textbf{90.12} & \textbf{90.38} & \textbf{88} \\
        \midrule    
    
        \multirow{8}{*}{Swin-S} 
         & \scshape{NBNet} & 
         26 & 29 & 37.12 & 31.75 & 32.75\cellcolor{red!25} & 29 & 30.57 \\
         & \scshape{LatentMap} & 
         16.12 & 14.37 & 14.37 & 5 & 21.62\cellcolor{red!25} & 13.75 & 12.72 \\
         & \scshape{Genetic} & 
         29 & 22.25 & 28.38 & 9.88 & 45.38\cellcolor{red!25} & 27 & 23.30 \\
         & \scshape{GaussBlob} & 
         4.62 & 29.25 & 7.12 & 4.62 & 17.25\cellcolor{red!25} & 4.5 & 10.02 \\
         & \scshape{EigenFace} & 
         26.88 & 39.38 & 36.5 & 25.5 & 55.38\cellcolor{red!25} & 30 & 31.65 \\
         & \scshape{FaceTI} & 
         46.5 & 47.88 & 53.25 & 43.5 & 63.75\cellcolor{red!25} & 50.5 & 48.33 \\
         & \scshape{QEZOGE} & 
         70.62 & 52 & 72.38 & 72.12 & 86.25\cellcolor{red!25} & 71.25 & 67.67 \\
         \rowcolor{Gray} \cellcolor{white} & \scshape{Ours} &  
         \textbf{90.38} & \textbf{76.38} & \textbf{89.5} & \textbf{91.62} & \textbf{90.5} & \textbf{89.62} & \textbf{87.5} \\
        \midrule

        \multirow{8}{*}{VGGNet19} 
         & \scshape{NBNet} & 
         62.75 & 22.25 & 72.12 & 42.12 & 22.25 & 72.12 & 44.30\cellcolor{red!25} \\
         & \scshape{LatentMap} & 
         26.75 & 16.25 & 22.75 & 5.75 & 18 & 29 & 17.90\cellcolor{red!25} \\
         & \scshape{Genetic} & 
         46.88 & 23.75 & 44.25 & 11.88 & 27.5 & 57.38 & 30.85\cellcolor{red!25} \\
         & \scshape{GaussBlob} & 
         5.75 & 20.38 & 6.38 & 0.88 & 1.75 & 7.88 & 7.03\cellcolor{red!25} \\
         & \scshape{EigenFace} & 
         87 & 36.88 & 83.75 & 59.75 & 60 & 91.62 & 65.48\cellcolor{red!25} \\
         & \scshape{FaceTI} & 
         9.12 & 7.38 & 10.38 & 1.5 & 7.12 & 12.5 & 7.10\cellcolor{red!25} \\
         & \scshape{QEZOGE} & 
         82.12 & 43.5 & 81.25 & 62.38 & 68.12 & 90.75 & 67.47\cellcolor{red!25} \\
         \rowcolor{Gray} \cellcolor{white} & \scshape{Ours} &  
         \textbf{93.5} & \textbf{61.12} & \textbf{91.38} & \textbf{84.88} & \textbf{85.12} & \textbf{93.5} & \textbf{83.20} \\     
         
    \bottomrule
    \end{tabular}
    }
\label{table:result_cfp_all}
\end{table*}

We conduct additional SAR evaluation on CFP-FP dataset with 800 pairs of frontal-profile pairs which is the maximum amount of positive pairs from the 200 generated frontal images in Tab.~\ref{table:result_cfp_all}. ALSUV shows the best performance comparing with previous baselines.

\clearpage

\subsection{Additional Baseline Comparison}

We additionally compare SAR results from LFW with GAFAR \cite{shahreza2023comprehensive} which is a blackbox method, and a whitebox method \cite{shahreza2022face} denoted as ICIP22. Our method surpasses both methods shown in Tab.~\ref{table:result_more_baseline}.

\begin{table*}[]
    \caption{Comparing with additional baselines with LFW dataset.}
    \centering
    \resizebox{0.97\linewidth}{!}{
\begin{tabular}{l|cccccc}
    \toprule
        &FaceNet & MobFaceNet & ResNet50 & Swin-S & VGGNet & AVG \\
    \midrule
        GAFAR\cite{shahreza2023comprehensive} & 20.9 & 26.42 & 23.64 & 14.17 & 27.62 & 22.55 \\
        ICIP22\cite{shahreza2022face} & 59.22 & 82 & 64.98 & 75.6 & 70.17 & 70.68 \\
        \textbf{Ours} & \textbf{99.36} & \textbf{96.16} & \textbf{99.56} & \textbf{99.93} & \textbf{99.56} & \textbf{98.91} \\
\bottomrule
\end{tabular}
    }
\label{table:result_more_baseline}
\end{table*}

\subsection{ALSUV in Blackbox Scenario}
\label{sec:alsuv_black_box}
ALSUV can be applied to blackbox scenario where gradients are not provided by combining with Zeroth-Order Gradeint Estimation(ZOGE). We compare with other blackbox methods in order to show the efficacy of ALSUV in blackbox scenario with results in Tab.~\ref{table:result_blackbox} with averaged results in unseen scenario. For training setup, we increase the total iteration to 300 with weight decay of 0.1 at iteration 150 with Adam optimizer. For hyperparameters, we use $n=20$, $t_{avg}=3$, and $k_{top}=5$.

\begin{table*}[]
    \caption{Comparing SAR of blackbox baselines and ALSUV generated under blackbox scenario where ALSUV is trained with Zeroth-order gradient estimation testing on LFW dataset. Results are the average of unseen scenario.}
    \centering
    \resizebox{0.97\linewidth}{!}{
    \begin{tabular}{l|cccccccc}
    \toprule
        Dataset & NBNet & Latent Map & Genetic & Gaussian Blob & Eigenface & FaceTI & QEZOGE & \textbf{black-box ALSUV} \\
    \midrule
        LFW & 22.21 & 11.40 & 27.21 & 9.97 & 57.21 & 18.73 & 69.49 & \textbf{82.38} \\
    \bottomrule
    \end{tabular}
    }
\label{table:result_blackbox}
\end{table*}

\clearpage
\section{Supplementary Setup}
\subsection{Face Recognition Encoders}
\label{sec:supp_setup}
Tab.~\ref{table:network_conf} shows the configuration of each face encoders used in all experiments. 
\begin{table*}
    \caption{
    Training configuration of the face encoders used in our experiments. For FaceNet, we use the pre-trained model from facenet-pytorch library. For MobileFaceNet and Swin-S, we use pre-trained models from \cite{wang2021facex}. For ResNet100, we use the model from the official ArcFace github. ResNet50 and VGG19 models were trained by ourselves. Finally, we use Swin-T for validation encoder. 
    }
    \centering
    \resizebox{0.7\linewidth}{!}{
    \begin{tabular}{lll}
    \toprule 
        Backbone & Classification Head & Dataset \\ 
        \midrule
        FaceNet\cite{schroff2015facenet} & Triplet Loss & CASIA-WebFace\cite{yi2014learning} \\ 
        MobileFaceNet\cite{chen2018mobilefacenets} & Additive Margin Softmax\cite{wang2020mis} & MS-CELEB-1M\cite{Guo2016MSCeleb1MAD} \\ 
        ResNet50\cite{he2016deep} & CosFace\cite{wang2018cosface} & MS-CELEB-1M \\ 
        ResNet100 & ArcFace\cite{deng2019arcface} & MS-CELEB-1M-v3 \\ 
        SwinTransformer-S\cite{liu2021swin} & Additive Margin Softmax & MS-CELEB-1M \\ 
        VGGNet\cite{simonyan2014very} & CosFace & MS-CELEB-1M \\ 
        
    \bottomrule
    \end{tabular}
}
\label{table:network_conf}
\end{table*}

\subsection{Dataset}
The description of the dataset used for our experiments are summarized in this section. For LFW and AgeDB-30, we compare reconstructed samples with every other positive samples except the one used for reconstruction, resulting in 3,166 and 3,307 pairs respectively. 
CFP-FP consists of frontal and profile images; we reconstruct only the frontal images and follow the challenging frontal-profile verification protocol, resulting in 130 pairs in total for results in Tab.~\ref{table:result_sar}. We conduct additional experiments in Supp.~\ref{sec:supp_results} Tab.~\ref{table:result_cfp_all} where we use all possible frontal-profile possible pairs resulting in 800 pairs for additional results. 
For identification, we set up generated samples as probes and every image as the gallery, resulting in 13,233, 2,000, and 5,298 samples in the gallery consisting of 5,749, 500, and 388 identities for LFW, CFP-FP, and AgeDB-30 respectively. For CFP-FP, we use generated frontal images for probes and profile images for the gallery considering the original design of the dataset.

\end{document}